# IMPLEMENTING EDGE BASED OBJECT DETECTION FOR MICROPLASTIC DEBRIS

ENV5555Y
By Amardeep Singh, Student #1005580300

## ACKNOWLEDGEMENT

The contents and results presented in this research report are the culmination of suggestions, tests, recommendations and improvements made over a long period of extensive tests to create salient solutions for pressing problems. I would like to extend my gratitude towards the Professor Charles Jia and Professor Donald Kirk for their constant support and constructive additions to the objectives of the project. A further extension of gratitude towards the School of Environment and the Faculty of Applied Science & Engineering for its extremely cooperative staff for allowing such research to occur in interdisciplinary vicinities. The research meetings and interactions helped in shaping the final objectives of this project and direct the simulations towards a more coherent path, where results can have a more meaningful interpretation among academics and researchers.

During the course of the previous months, several new problems and issues had emerged which wouldn't have been resolved without the continued insights from the Faculty and the research staff. The flexibility of approach and the independence to build solutions using new ideas truly assisted in the development of the solutions discussed in this paper. The simulations and results have been built using open source and paid software with a proper focus on creating models that can be replicated and used by others. The far larger application of these results would be integrating the models with hardware devices for the purpose of identification, segregation, tracking and analysis. Particular attention has been devoted to making comparisons for this in the concluding sections, none of which would have been possible without support and use of proprietary virtual software from Intel and its platforms. I am highly appreciative towards the University Of Toronto for creating such a collaborative experience that equips students with resources to conduct research and communicate ideas with a large network of professionals and experts.

The methods and results discussed in this research report serve as a foundational point for those looking to create feasible solutions for tackling waste problems, especially in the context of plastic debris which requires action from various facets for resolution. Finally, I would also like to express recognition of support from the vast network of researchers, datasets and research papers that played a decisive role in developing new algorithms and frameworks for meeting objectives.

# KEY TO ABBREVIATIONS:-

| | |
|---|---|
| **Adam** | Adaptive Moment Estimation |
| **AI** | Artificial Intelligence |
| **Autocaffe** | Automatic Convolutional Architecture for Fast Feature Embedding |
| **AWS** | Amazon Web Services |
| **CCTV** | Closed-circuit television |
| **CNN** | Convolutional Neural Networks |
| **COBWEB** | Complexity and Organized Behaviour Within Environmental Bounds |
| **COCO** | Common Objects in Context |
| **CPU** | Central Processing Unit |
| **ESG** | Environmental, social and corporate governance |
| **FFMPEG** | Fast Forward MPEG |
| **FFSERVER** | Fast Forward Server |
| **FPGA** | Field Programmable Gate Arrays |
| **FPS** | Frames Per Second |
| **GIS** | Geographic information system |
| **GPU** | Graphics processing unit |
| **HOG** | Histogram of oriented gradients |
| **IGPU** | Integrated Graphics Processing Unit |
| **IoT** | Internet Of Things |
| **IR** | Information Retrieval |
| **KNN** | K Nearest Neighbors |
| **LCA** | Life Cycle Assessment |
| **MQTT** | Message Queuing Telemetry Transport |
| **MTS** | Marina Trash Skimmer |
| **MXNet** | Mix and Maximize Network |
| **NCS2** | Neural Compute Stick 2 |
| **NGO** | Non-Governmental Organization |
| **NOAA** | National Oceanic and Atmospheric Administration |
| ONNX | Open Neural Network Exchange |
| OpenCV | Open Computer Vision |
| ReLu | Rectified linear activation function |
| RFID | Radio-frequency identification |
| SGD | Stochastic Gradient Descent |
| SIFT | Scale-invariant feature transform |
| SVM | Support Vector Machines |
| TDP | Thermal Design Power |
| USB | Universal Serial Bus |
| VPU | Vision Processing Unit |
| WTE | Waste to energy |
| XGBoosting | Extreme Gradient Boosting |

Note:- Images that have been used from internet and other research sources have been referenced under in the figure captions. Images that appear without a caption in the reference were obtained from the simulations and experiments conducted on coding platforms. Others have been collected from the results of algorithms executed in cloud environments.

# 1. ABSTRACT

Plastic has imbibed itself as an indispensable part of our day to day activities, becoming a source of problems due to its non-biodegradable nature and cheaper production prices. With these problems, comes the challenge of mitigating and responding to the aftereffects of disposal or the lack of proper disposal which leads to waste concentrating in locations and disturbing ecosystems for both plants and animals. As plastic debris levels continue to rise with the accumulation of waste in garbage patches in landfills and more hazardously in natural water bodies, swift action is necessary to plug or cease this flow. While manual sorting operations and detection can offer a solution, they can be augmented using highly advanced computer imagery linked with robotic appendages for removing wastes [1]. A common problem that occurs with floating debris is that such robotic machines can cause harm to fish and other marine life due to failures in detection. The purpose of this research project is to investigate some general frameworks that have been built on existing platforms for object detection and improve upon them by building the models around wastes, leading to refined plastic waste detection models. These models can thus be integrated with ocean sweepers like the MTS (Marina Trash Skimmer) or the Seabin (commonly popular in harbors and rivers globally) for better waste profiling and removal. A future rendition of the machines that are capable of underwater mobility could utilize the developed models for removal with minimal human intervention, acting like clean-up vehicles, sweeping away the ocean floors for waste. The primary application of focus in this report are the much-discussed Computer Vision and Open Vision which have gained novelty for their light dependence on internet and ability to relay information in remote areas. These applications can be applied to the creation of edge-based mobility devices that can as a counter to the growing problem of plastic debris in oceans and rivers, demanding little connectivity and still offering the same results with reasonably timed maintenance. The principal findings of this project cover the various methods that were tested and deployed to detect waste in images, as well as comparing them against different waste types. The project has been able to produce workable models that can perform on time detection of sampled images using an augmented CNN approach. Latter portions of the project have also achieved a better interpretation of the necessary preprocessing steps required to arrive at the best accuracies, including the best hardware for expanding waste detection studies to larger environments. These models and tests have been validated using standard procedures that are often employed in machine learning projects, which testifies their position as potential placeholders to facilitate microplastic detection and cleanup.

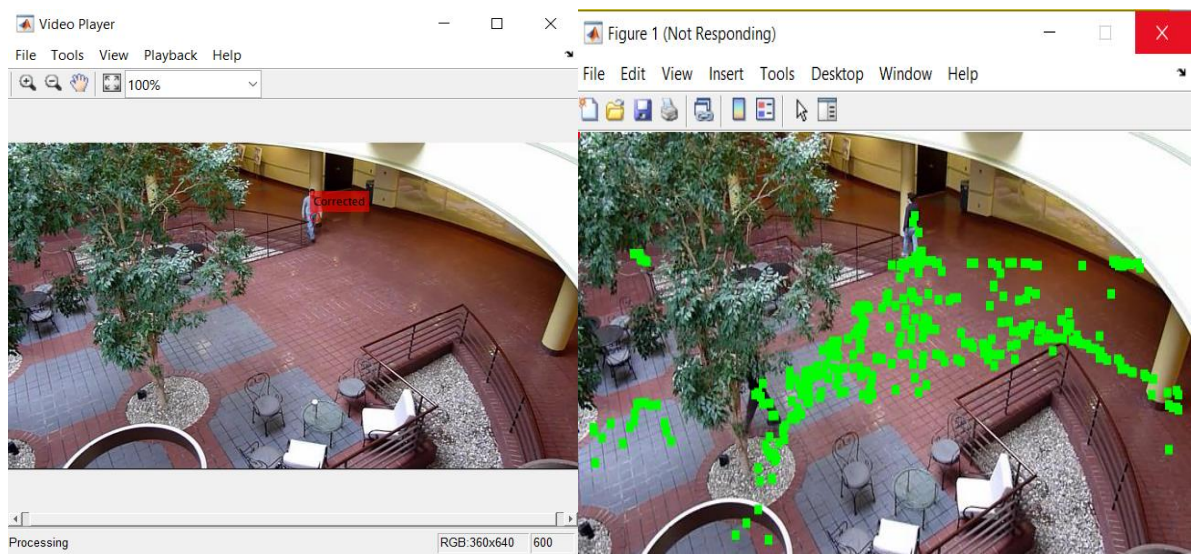

*Figure 1.1. Tracking software results for person detection in MATLAB using the Computer Vision Toolbox. The Corrected tag and the green squares are the algorithm's results of tracking a moving person in the frame.*

# 2. INTRODUCTION- THE PLASTIC PROBLEM

Waste outputs and volumes are a pressing issue for nations as the human footprint on the planet continues to take a toll on the environment. Statistics have consistently shown that human generated waste has increased monumentally with the rapid development of urban areas, industries and human settlements which has produced a multitude of solid emissions including paper, plastic, wood, rubber, glass and so on [2]. Due to the non-biodegradable nature of a majority of such types of wastes, it becomes a challenge for communities to find ways to reduce their impact while devising new methods for reusing and recycling. An example of such impact can be understood from ocean garbage patches, primarily consisting of plastic waste, that have accumulated at various locations globally to hamper marine ecosystems. The Great Pacific Garbage Patch, for example, covers approximately 1.6 million square kilometres and continues to grow due to the action of ocean currents and unabated human waste disposal. As a response to growing plastic wastes in the oceans, non-profits and startups have been set up which use both mechanical and robotic means to extract the waste. Technologies such as the Marina Trash Skimmer and the Industries and communities in several Asian countries have developed a pipeline to connect waste hotspots with a supply chain mechanism to direct them to the appropriate location, where it can be processed and reused. However, a primary challenge in the process of waste collection and segregation is the time taken to separate materials by material type to reduce their final environmental impact.

If we take the statistics into a glance, we can see that global primary production of plastic was nearly 270 million tonnes in the year 2018 with coastal regions being at the biggest risk of turmoil. In 2010, coastal plastic waste – generated within 50 kilometres of the coastline – amounted to 99.5 million tonnes with improperly managed (mismanaged) waste contributing to the leakage of more than 31.9 million tonnes of waste [3]. Comprising of this, 8 million tonnes of waste has already been added to the oceans. Plastic in oceans presents a multitude of challenges, not just restricted to the floating patches but also the floating pieces that can dissociate and affect ecosystems for longer periods. To this end, total amount of plastic in surface waters is still unknown, with normal estimates being between 10,000s to 100,000s tonnes.

Human dependency of plastic can be seen as the main culprit fueling this wanton release of waste. The first synthetic plastic that was released to the public was Bakelite in 1907 which would usher into the creation of the global plastic industry by the 1950's, raising the global level by 200 fold to 381 million tonnes in 2015. While solutions for creating biodegradable plastic may pose a counter to this, the longevity of total degradation complicates the cradle to grave design for materials. More importantly, it can be ascertained that poor management of waste is far more negative in plastic waste release which ultimately overwhelms any amount of biodegradable volumes of plastic. Current statistics show that developed and developing nations both are equally playing a role in this release with an estimated 20 percent of all plastic waste in the oceans originating from marine sources. A major portion of the wastes in the Great Pacific Garbage Patch (GPGP) originate from fishing nets, ropes, and lines. To contrast, plastic levels in production reached 2 million tonnes per year in 1950, seeing dips and turns depending on global financial profiles and the pricing of petroleum.

Moving on to the industry's response to this problem, one can see several solutions being tested out. Prior to 1980, recycling and incineration of plastic was seen to be negligible and all plastic was discarded as a result [4]. With tighter regulations on the emissions produced from these wastes, the interest has shifted from total disposal to incineration, averaging by about 0.7 percent per year since 1980. In 2015, an estimated 55 percent of global plastic waste was discarded, 25 percent was incinerated, and 20 percent recycled. An extrapolation of these trends indicates that nearly half of all plastic waste is likely to be incinerated by 2050 while 44% would be restricted to recycling and the remaining would be discarded. This is, again, a highly conditional assumption.

To summarize, the major objectives of this study were to create a feasible batch of images for different kinds of waste(mainly consisting of municipal solid wastes and images obtained from microplastic sources), testing out the images in a collection of different algorithms and configuring the best choice for experimental research studies. The project experiments and algorithmic tests include conventional approaches but evolved into ensemble-based detection schemes built on convolutional neural networks. This forms another significant aspect of the design study where a great deal of time was spent on comparing different activation functions for analyzing imaging data in neural networks. The project intended to create a system of preprocessing steps (as all steps are not useful) by changing image dimensions, depth, colors and studying the effects of altering capturing data (focal length, distancing) on the prediction algorithms. Finally, by taking reference from existing sources for video detection and hardware, the project also looked into comparing hardware requirements for different use cases which will help in the development of better testing frameworks to collect and analyze microplastic debris in water bodies.

## 3. PLASTIC WASTE AND ITS IMPACT

When glancing at the information that we have about the waste floating and dispersing in the global water bodies, one can see that polystyrene foam (Styrofoam) and its derivatives comprise 90% of all marine debris, all of it falling under the single-use food and beverage category. Due to the action of ocean currents, plastic waste breaks down into more hazardous microplastics or beads that has a far greater longevity. Life cycle assessments of plastic wastes have confirmed that plastic waste has been able to reach other portions of the biosphere including the soil, the sea floor, terrestrial animals and even human food. The typical size of a primary microplastic bead (estimated to number close to 5.5 trillion individual pieces in the oceans) is considered to be 5.0 mm and dimensions can be even smaller for pieces that have experienced degradation by natural agents.

Surveys and research have proven that microplastics affect the quality of life on marine ecosystems as well as human communities who depend on the water bodies for their subsistence and livelihood. Polystyrene beads are soft targets that are not easily digested and can thus accumulate in the digestive tract of marine animals who consume them [5].

Microplastics also arise from clothing and textiles, especially non-organic materials like rayon or nylon which closely mimic polystyrene in its properties when released into the seas. Moreover, due to their minute size, these fibres can easily pass through treatment units in wastewater plants and find their way into oceans. Microfibres have been found in many different ecosystems, including freshwater systems, ocean waters, ocean sediments, and beaches around the world, indicating it is a worldwide problem that is possibly growing. This makes textiles the second biggest source of microfibres and microplastics along with food and beverage materials. Materials discarded from car tyres account for 28% of the total discarded waste, making them the third largest type of waste. Plastic packaging is lightweight, causing it to be carried over by the wind or washed into rivers, where it reaches the sea.

## 4. MITIGATING PLASTIC WASTE

One of the most common methods of clearing up wastes that have found their way into water bodies, is by organizing cleanup efforts, led by NGOs and global environmental preservation groups. While beach cleanups and setting up separatory units at the openings of water bodies like riverbanks and river mouths, such methods take vast amounts of human participation and long time to review the cleanup.

Groups like The Ocean Cleanup Project and NOAA often take use from volunteers to organize cleanup drives and account for the collected waste using common methods of analysis. There has been a greater push to invest in more sophisticated technology for the detection and classification of waste using robotic cleanup machines that can scour the sea surfaces for collecting waste. A bigger challenge however remains waste that floats about in deeper ends and poses threats to natural ecosystems. Other methods of ceasing the flow of plastic debris into the oceans takes both a preventive and reactive approach [6]. Preventive efforts include minimizing the use of single use plastics and utilizing alternatives for commonly available plastic items including brushes, bottles, bags and clothing. Industries too, under the scrutiny of regulations, have ramped up proper segregation principles such as Kiverco which has already pioneered a semi automated method to deal with municipal waste.

The total waste coverage that cleanups can take up is still highly debatable and would require consistent effort over longer periods which makes them difficult to organize and execute. In this regard, machine automated methods can prove to be a boon but require constant maintenance and some level of human control to know where to search and collect. Ocean cleanups and beach cleanups also lack the access to far off locations, as volunteers mostly depend on ocean and sea currents to wash away the waste on shore and on the coastal regions, to allow for picking up. Stationary units that derive much of their waste collection by remaining in single locations and filtering out solids with moving currents, cause problems as marine life can get caught in the machinery (turbines and rotor blades) producing new challenges. The alternative to this would be the use of simple nets that capture the waste depending on the pore size of the materials but are again dependent on the current motions to truly collect it.

# 5. SUSTAINABILITY & IMPACT

Modelling waste streams and their effects have been conducted on several softwares in the past but to truly incorporate the elements of an interactive society is highly complex. Current methods of waste management involve two major operations: -
1. Dilute and disperse.
2. Concentrate and contain.

These methods are more influenced on the types of solid wastes that enter the pipeline which can be best classified by the portion that they occupy in the product to disposal cycle: -

-Wastes can be classified based on the source and type of material utilized in composing the primary materials. The broader definition of biodegradable and non-biodegradable wastes can be extended to include plastics, metals, non-metals and composites. Industrial terminologies for this can include materials and wastes that are further subdivided as perishables and non-perishables. Disposal of the wastes can differ depending on the end goal of the process- either to dilute and disperse or concentrate and contain. Methods can be differentiated as primary, secondary or tertiary: -

-Primary Waste Disposal:- Wastes are disposed in a single container or spot but are not separated into sections for more homogeneous partitions. The purpose with primary waste disposal is to merely collect and disperse waste materials in landfills, encampments, and bins. The rate of mixing and heterogeneous makeup of materials makes it a capital intensive process later on.

-Secondary Waste Disposal:- More advanced than primary methods of waste disposal that can involve manual picking, separation, composting etc. Secondary waste disposal helps in delivering wastes to more useful locations where they may be processed or reduced to reduce the implications of environmental degradation.

-Tertiary Waste Disposal:- Tertiary waste disposal methods include more specific waste utilization methods such as incineration, biodegradation, chemical breakdown etc. Wastes tend to be classified into greater degrees of separation including toxicity, and chemical reactivity.

As more and more companies adopt better ESG standards with an intention to improve their performance and general public outlook, waste management technologies become more visible. The Global industrial solid waste management market is already set to surpass USD 1.1 Trillion by 2026, as reported in the latest study by Global Market Insights driven by an urge to minimize adverse effects of untreated industrial waste on public health and environment [7]. Government regulations too have a great role in driving this sustainability agenda and creating new businesses that focus on completing the 'cradle to grave' philosophy.

Global industrial solid waste management industry across mining industry will grow owing to large amount of waste generation and adverse effects on workers and the surrounding ecosystem. Moreover, introduction of stringent regulations pertaining to effective treatment of the generated waste to curb the degradation of mine workers' health will drive the technology deployment. In addition, demand for suitable treatment technologies during mineral ore extraction & recovery to minimize the generation of hazardous gases will further favour the global industrial solid waste management market growth.

Open dumping is the most practiced solid waste treatment process across the low- and middle-income economies owing to the lack of availability of effective disposal techniques. However, proactive policy measures to minimize the impact of industrial waste streams on environment is set to decline the open dumping over the forecast timeframe. Moreover, declining technology costs and product innovations has created an availability of better disposal methods, which will reduce the open dumping practices in forthcoming years.

Based on several statistical reports highlighting the future of waste and its impact on the environment and global sustainability as a whole, one can see the following trends:-

1. Rapid industrialization across developing economies along with efforts to attain a sustainable growth will produce a demand for effective waste management practices. It is anticipated that unlike western nations, those in the developing world will see an influx of new startups led by the private sector which will essentially merge into cooperatives with time.

2. Stringent government policies to reduce the environmental impact of industrial solid waste will boost the business growth in a few select nations. As carbon pricing and sustainable finance becomes a norm pitting big polluters against each other, an artificial economy will be created much like the one emerging after the introduction of carbon credit trading in European countries.

3. While new startups and companies will take shape, they will eventually roll into larger corporations. Some of the big players in the future in terms of waste management will include the Biffa Group, Hitachi Zosen Corporation, Green Conversion Systems, Covanta Holding Corporation, Plasco Conversion Technologies, Veolia, SUEZ, among others.

4. There will be a growing emphasis on circular economies and material recycling as reusing materials will dictate the growth of the waste management market [8].

5. Solid waste management markets across textile industries will witness growth owing to rising clothing materials demand fuelled by expanding population and improving standards of living. In addition, reusability of fabrics will prompt recycling technologies across textile industry, thereby augmenting the industrial solid waste management market.

6. Industrial solid waste management in the Latin American market will witness growth on account of ongoing industrialization, already having added USD 755.4 billion in 2018 from USD 627.6 billion in 2009. Investments in infrastructure would have to continue unabated to accomplish this.

7. The food sector industry will also see great improvements in waste management technology. To combat the rising levels of food waste, systems dynamic modelling softwares would be integrated with better preservation techniques and perishable models to accurately transfer edibles before they spoil. Monitoring technologies can extensively enhance the performance of the food supply chain decreasing product loss, especially through computer vision based models which are discussed in the future sections.

8. Inventory operations are imperative because they manage the material flows in highly alternating conditions. The deterioration level of the products as well as the market demand have a strong effect on warehousing strategies.

9. There will be a greater acceptance of academic backed methods such as Monte Carlo methods to develop better probabilistic models, combined with GIS technologies to properly navigate the best possible locations for disposal.

10. Rather than having operations managers work using singular models to arrive at conclusions about the best methods of execution, softwares and tools will adapt faster to constraints such as regulations and pricing of recycled materials, in a manner similar to the business information management tools used extensively in architecture. System dynamics methodology is another scenario in research that has been fantastic in handling specific waste administration issues.

11. Enhanced process control methods will move beyond the conventional decision making algorithms to include more neuro and neuro-fuzzy approaches such as in forecasting different scenarios and predicting environmental impact.

## 6. **WASTE GENERATION- GLOBAL STATISTICS**

Globally, waste emissions have been at an all time high due to a variety of factors. The year 2016 alone exhibited levels as high as 2.01 billion tonnes of solid waste, amounting to a footprint of 0.74 kilograms per person per day. From here on, waste levels are expected to rise by 70% from 2016 levels to 3.40 billion tonnes in 2050 [9].

The World Bank and joint projects with the World Energy Council show that compared to those in developed nations, residents in underdeveloped nations are more impacted by unsustainably managed waste. In low-income countries, most waste ends up in unregulated dumps or is openly burned leading to serious health, safety, and environmental repercussions. Poorly managed waste becomes a vector for diseases, fastens climate change factors through methane generation and cause great disturbances in socioeconomic conditions. Recent controversies over the transfer of rusted ships to South East Asian countries to be converted to scrap metal is one such example.

One of the major challenges that developing nations face in terms of effective waste management is the expense associated with sound mitigation strategies, as most industries can allot 20%–50% of municipal budgets. Operating this essential municipal service requires integrated systems that are efficient, sustainable, and socially supported. The World Bank alone has invested around $4.7 billion since 2000 to more than 340 solid waste management programs.

Developed nations like the U.S. and Canada represent just 16% of global population, but house industries that generate nearly 34% of the world's waste. An estimated 93% of waste in low-income countries is mismanaged, as compared to 2% in high-income countries. Estimates indicate that the daily per capita of waste is sitting at 4.87 pounds per person in North America as compared to 1.01 pounds in Sub-Saharan Africa. The rate of increase in both these regions will account for 35% of global output by 2050.

Collection is more common in urban areas for these low-income countries than in rural areas, but still not as prevalent in developed nations. The World Bank found that the amount of collection in low-income countries increased from about 22% to 39% since 2012 — though noted data may not be directly comparable. Waste emissions have added 1.6 billion metric tons of carbon-dioxide-equivalent since 2000 and is expected to grow to 2.6 billion metric tons by 2050 [10].

Alarmingly, food waste accounts for 47% of these emissions mainly caused by problems of supply chain, storage and inadequate scheduling. Underdeveloped waste management infrastructure may also become a growing sign of global inequity as climate change advances. There is also a great disparity in the options available to deal with the waste including the Incineration of waste for conversion to energy which is more widespread in wealthier nations. It's believed that about 22% of waste is handled at waste to energy (WTE) facilities.

The World Bank estimates the use of WTE in upper-middle-income countries has increased from 0.1% to 10% since 2012 with more capital being funnelled into research and development and a sizeable amount being compartmentalized for subsidized processing units.

## 7. A NOTE ON AUTOMATED METHODS

Some of the more noted technologies in the field of automation that are seeing extensive attention in the field of waste management include the following:-

1. Automated Waste Sorting

Companies like Kiverco excel in combining both manual and automated sorting methods. Another example of an automated sorting giant is the Finnish company, ZenRobotics, that uses artificial intelligence for smart recycling by handling waste materials with the use of an automated waste sorter. Most sorting plants use the conglomeration of machine learning, artificial intelligence, and computer vision, combined with robotic appendages that pick out and classify the wastes.

2. RFID Tagging

While RFID tags are seen as a novelty more famous among nuclear industries for tracking and keeping a close watch on the movement of nuclear waste, RFID tagging methods have caught on as important additions to vehicles carrying waste, LCA studies, containment devices such as hoppers, silos and so on. Waste materials picking or sorting structure has seen a change in the advancement of RFID tags.

Songdo, a South Korean city, for example, uses RFID tags to identify the trash, under different tags. These tags are then fed into a waste disposal system which scans the tags and directs them to different locations depending on their final use, biodegradability and material of construction. It is mainly driven by a single server who closely surveys the flow of waste.

3. IoT Sensors and Artificial Intelligence Programs

IoT(Internet of Things) encompasses all items, applications and technology that connects devices to the internet and allows users to utilize this interaction to make automated changes. IoT currently exists for several household items including television, refrigerators, thermostats and security systems. The concept of interactivity between devices and users is heavily dependent on how many activities can be performed using short range signals from a remote controlled device. IoT devices receive updates and changes through a local network by the user and can be automated to perform special tasks.

The intelligent trash can has been a conceptualized idea for a smart disposal container that can take in different sorts of waste. Built on a framework of artificial intelligence software programs and IoT detectors, face roadblocks in adoption due to their pricing and maintenance. Present day sensors and AI programs measure the waste thrown inside them and send this information via advanced servers to the major removal system for processing.

The system at the backend then uses this information to categorize the waste based on type, level of environmental impact and quantity, directing it to different waste facilities. Current frameworks like OpenVino from Intel can offer great implementations in the future to bring hundreds of USB sized units for detection in industries.

4. Computer Vision Programs

Open Vision is another terminology for System Vision and Computer Vision that describes the methods, technology and algorithms that involve the detection and determination of objects of interest using automated means. Open Vision is directly related to modern platforms such as OpenCV, TensorFlow, Python, Caffe, and ONNX. It is commonly used for building applications in devices for tracking and detection and has gained greater popularity in self driving vehicles for detecting objects in frames while in motion. Open Vision concepts can be modified to be applied to smaller machines for longer periods including home cameras, computer cameras and other devices that can be connected to the internet.

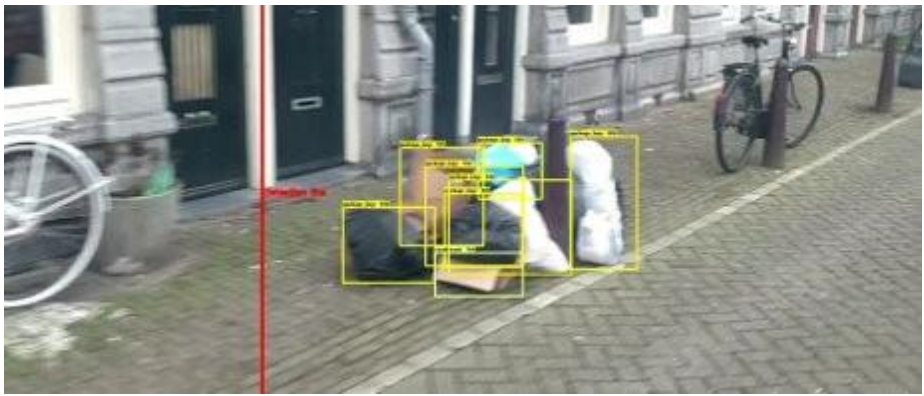

*Figure 7.1. Street Trash detected using Bounding Box Algorithms(Source:-Code For NL)*

While manual picking is still the default choice for industries to segregate and control the flow of waste, it is often tedious, slow and prone to errors. Modern technology has allowed the addition of detection systems using close circuit video feeds linked with robotic appendages to detect and pick out the waste as it moves on the pipeline. More advanced trash dumps have begun using waste picking and sorting robots.

5. Using Artificial Intelligence for Smart Waste Management

Often seen as the bells and whistles due to its ubiquitous nature to be useful anywhere and everywhere, AI systems for smart waste management are used mainly to deal with electronic waste and plastic debris in oceans. Global plastics manufacturing had increased from 250 million metric tons in 2017 to 350 metric tons in 2018, with only 15 percent of this being recycled wither due to poor segregation or mismanagement of waste streams. The other 85 percent is either trashed into landfills or penetrates the environment [11]. One example of a system to combat this is the SFU Mechatronics Systems that created an AI-powered smart waste management equipment to that automatically sorts garbage.

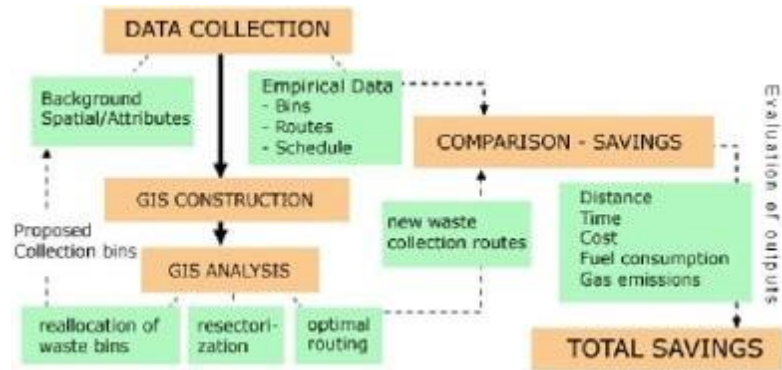

*Figure 7.2. AI methods for GIS determination of disposal sites [12].*

6. Systems Dynamics Modelling

A form of interactive modelling that is used by researchers to simulate realistic social cycles of production and emissions. Systems dynamics modelling uses equations derived from statistical data and allows them to be applied to user defined input output frames. It is an approach to understanding the nonlinear behaviour of complex systems over time by using stocks, flows, internal feedback loops, table functions and time delays. In the context of this project, it can be used to develop models for waste generation to understand what are the main factors affecting its trends and impact in the environment, by connecting them to anthropogenic(human) inputs such as daily waste generation, number of nearby industries etc. Softwares like Vensim and COBWEB are good examples of platforms used for systems dynamics modelling.

7. Edge Applications

Edge applications are an extension of wireless devices that grew in use due to the popularity of edge computing which uses advanced methods for storing, handling, processing and delivering data to devices around the world. Edge applications are characterised by their small real-time computing power requirement and multipurpose features. Most edge applications are connected to cloud infrastructures and require minimal intervention by users for setup and maintenance. Data computation and data storage can be performed faster on edge applications even from larger distances on strong networks like 4G. Real time data can also be processed locally to reduce latency and cut down on the capital required to run the machines.

# 8. COMPUTER VISION AND WASTE

Computer Vision is the application of a computerized interface or platform for executing algorithms that can detect, classify and recognize objects, instances and even people. The current application of computer vision is well known in automotive sectors as well as industrial workshops where surveillance of surroundings is highly important. The typical computer vision system uses a system of cameras and sensors that relay the visual data back to a computer system where it is preprocessed and analyzed to show desired results. Depending on the quality of the feed, the hardware used, the detected object of interest and the preprocessing that is done to create the models. For applications that use cameras as their sensors, there is a heavy reliance on images to for collecting information about the objects to be detected. More specifically, they used computer vision to help them identify objects within an image. Computer vision is a computer science concept that enables a computer to identify objects autonomously. Computer Vision is more commonly famous for being the staple tool for companies like Tesla, Nvidia and Uber for autonomous driving systems. Most of these applications use frameworks for grayscaling, thresholding, SIFT and KNN algorithms.

A subsection of computer vision involves some image processing techniques in order to alter the image. Image processing is very important for extracting features within an image as it makes it easier for computers to differentiate concrete objects from noise. However, in some cases, it can become important to add noise in order to make the model for powerful if the object of interest is really hard to distinguish. Some basic image processing methods include grayscale and thresholding. These are important as they help in the removal of unnecessary features from images and can isolate the object from the surrounding. Feature extraction can involve the separation of details such as dimensions, color, shape, pixel density and more. Image processing techniques are important to help connect the link between the output and the input, but due to the black box nature of models, it can become difficult to interpret which features are far more important. This can however be remedied using conventional semi supervised models. Such methods are necessary in the identification of objects and creating comparison matrices that can help in clustering the various items in an image. Some the methods such as SIFT and Hu involve more advanced methods such as a scale invariant transformations and histogram of oriented gradients(HOG) that uses ta sliding window of detection around the image until a reasonably congruent object id is detected. At every position of the detector window, HOG descriptors are computed for the detection window.

The feature extraction step is followed by classification which can be accomplished using multiple methods. The classification methods combine the extracted features along with the ground truth to classify objects. The project in this context used the conventional models from ImageNet, OpenVino, ONNX and COCO to test out the accuracies of using object detection mechanism and then moves to more adaptive CNN based architectures to help in removing the inconsistencies of using the former. The methods that are used for algorithmic detection can include the typical k-nearest neighbor (KNN) algorithm, the support vector machine (SVM) algorithm and convolutional neural networks (CNN). For the purpose of this project, the waste debris was detected initially using a sequence of images and in a batch and then evolved to study waste in floating environments captured through videos and robotic drones. Processing of images/videos was performed using simple pre-processing steps that improved on corrections in the color, density, hue, saturation and other media-based elements.

All of the methodologies (sensors and computer vision algorithms) that were discussed above need a computational device that can serve as the platform for directing video or image feed to setups and then running the algorithms onto them. For this purpose, the project used a CPU environment Core i7(8$^{th}$ Gen) from Intel. Further studies on hardware were accomplished using more streamlined comparisons using the OpenVino environment and by running the coded algorithm on batches using AWS (Amazon Web Services). There are however other hardwares and softwares on the market which are beyond the scope of this project. As an example, capacitive or light sensors can be employed in conjunction with a microcontroller which may be the only computational device necessary for the system whereas a convolutional neural network (CNN) may require more computational power. Smaller computing units like Raspberry Pi's and Arudinios have been known to accomplish object detection using similar methods but cannot be expanded to handle the demands of a hugely continuous data feed. Such units utilize a computer with a high clock speed for its central processing unit (CPU) and a dedicated graphics processing unit (GPU) with several gigabytes of RAM.

# 9. THE FUNDAMENTALS OF OBJECT DETECTION

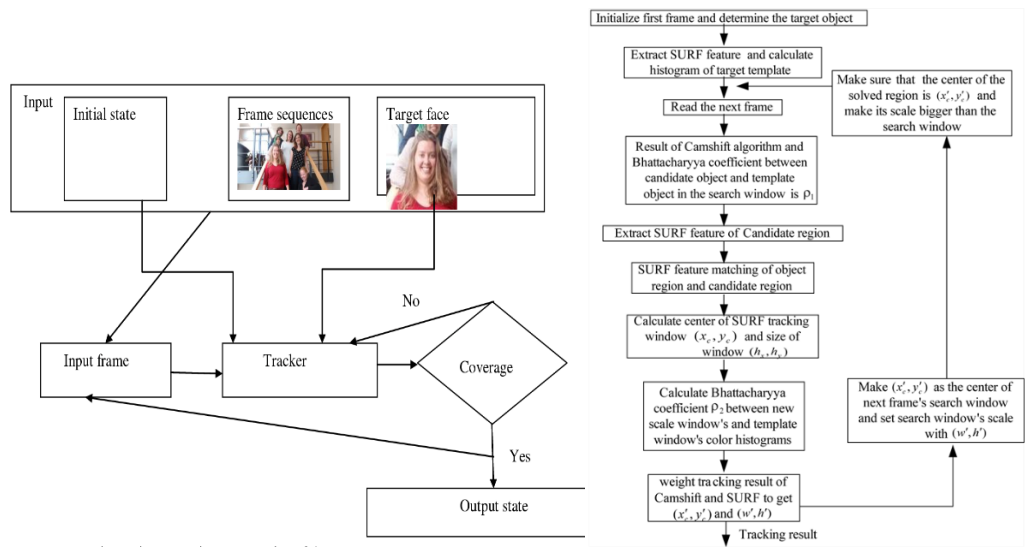

*Figure 9.1. General schematic of a tracking algorithm and preprocessing system. The algorithm breaks down the frames and compares the input features against a coverage(ground truth). The image on the right is an example of the CAMShift(Continuously Adaptive Mean Shift) algorithm used for object tracking. [13]*

Object detection for waste has been well established for treating municipal waste. Open vision algorithms openly contribute to the circular economy through the recovery of recyclable materials from the waste stream. Despite the significant efforts made by local governments to implement waste segregation at the point of collection, large amounts of recyclables still find their way into the incorrect streams. These applications recently have caught the interest of researchers for treating plastic debris, and to an extension, to microplastics.

Waste treatment facilities can recover a large percentage of materials but must deals with a highly variable input stream that can limit its ability to reach 100% accuracy for detection. This is due to current processing methods having a clear lack of featured characteristics that can make the models more adaptable. Real time monitoring can help achieve this which when combined with different architectures for CNNs, can make detection and tracking more efficient. Under current conditions, obtaining valuable data (such as traceability of materials, mass balance and performance of the separation equipment) is a strenuous task due to the large volumes of waste. Other discrepancies that affect the modelling arise from the composition of the the waste and equipment performance deviations. Examples of companies that are working to counter this trend are startups like Sadako, which is using a combination of AI and computer vision to build better decision-making algorithms. Current research suggests that multilayer neural networks are better at reaching the goal of better accuracies. As more detection systems are setup in waste management facilities and more datasets are generated, better models can be made available for public and academic use. Present algorithms insert an array of monitoring points, installed throughout the separation process, which are then linked to a central server that collect all input feed. This is attached to a backend where the processing occurs, and the results are generated for comparison.

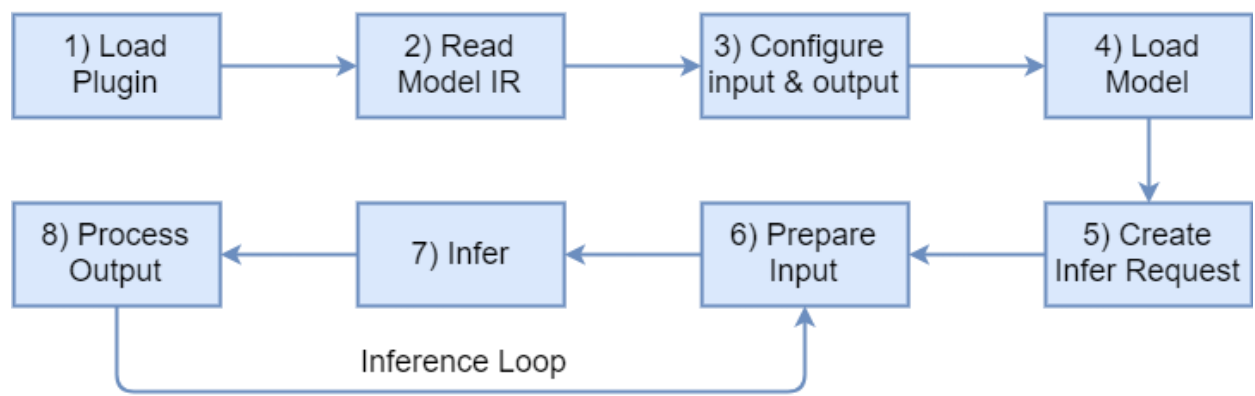

*Figure 9.2. Steps involved in using a pre built model from loading plugins to processing inputs and updating the inputs with inference loops.*

A MQTT server was used for creating sample calls to the OpenVino platform. A telemetry transport uses a normal publish/subscribe architecture that is designed for devices lacking specific resources and low-bandwidth setups, which is used extensively in IoT devices. It uses a publish/subscribe architecture that is composed of a broker, or hub, that receives messages published to it by different clients. The broker then routes the messages to any clients subscribing to those particular messages.

This is managed through the use of elements called topics. As one client publishes to a topic, while another client subscribes to the topic. The broker handles the conveyance of this message from the publishing client on that topic to any subscribers. These clients therefore don't need to know anything about each other, but the topics which they want to publish or subscribe to. MQTT is one example of this type of architecture, and is very lightweight, offering great capabilities for use in realistic projects. The publish/subscribe mainframe is also used with self-driving cars, such as with the Robot Operating System, or ROS for short. This uses a stop light classifier that may publish on one topic, with an intermediate system that determines when to brake subscribing to that topic, and then that system could publish to another topic that the actual brake system itself subscribes to. In a publish-subscribe architecture, the entity that is sending data to a broker on a certain "topic" is the publisher or sender while the entity that is listening to data on a certain "topic" from a broker is called the receiver or subscriber. In a publish-subscribe architecture, data is published to a given topic, and subscribers to that topic can then receive that data. It is important to understand that the at most fundamental level, the processing and analysis of videos is performed in the same manner as any other numeric machine learning. As a result, high end data found in images and videos must be converted into workable formats which can be performed using ffserver softwares. These softwares stream to a web server, which can then be queried by a Node server for viewing in a web browser. Current age softwares that are used in municipal waste management use a web server built on a Node.js to handle HTTP requests and/or serve up a webpage for viewing in a browser.

# 10. COMPUTER VISION AND IMAGE DETECTION

The project was initially started with an objective of detecting plastic debris alone using conventional computer vision algorithms that would employ either bounding box or image segmentation steps to produce results. However, as the project proceeded, several challenges were observed in the general detection of waste, due to the model's poor ability to differentiate between different waste pieces and the surroundings. The flowcharts below show the schematic for two methods that were employed to achieve this result. The project initially looked into training images using a bounding box and segmentation algorithm which were processed using tesseract. These images were then reshaped and resized to be made feasible for analysis in a simple 3x3 neural network framework. The final accuracy of the algorithms was compared by measuring the number of measured boxes against the actual pieces of waste debris in a photo. Using a small batch of 300 images, several problems and challenges emerged.

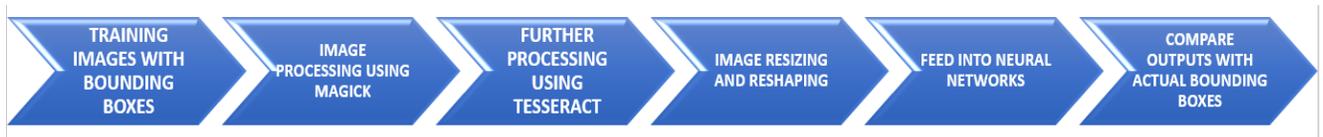

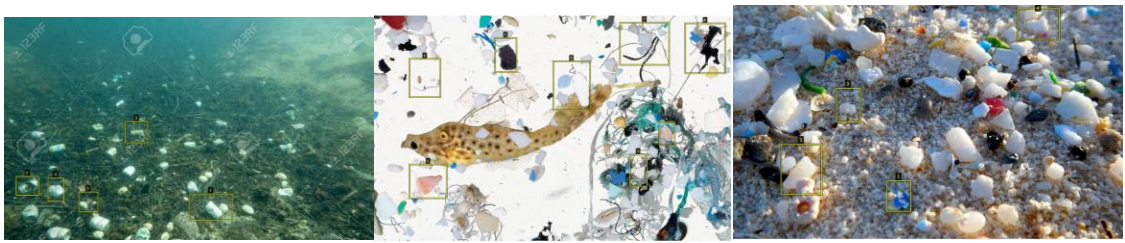

*Figure 10.1. Pipeline steps for object detection for debris using the magick and tesseract package. The images below the steps are the bounding box results for a sample of images where the yellow square boxes indicate the algorithm learning information about the debris.*

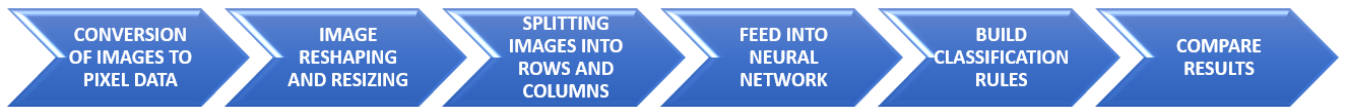

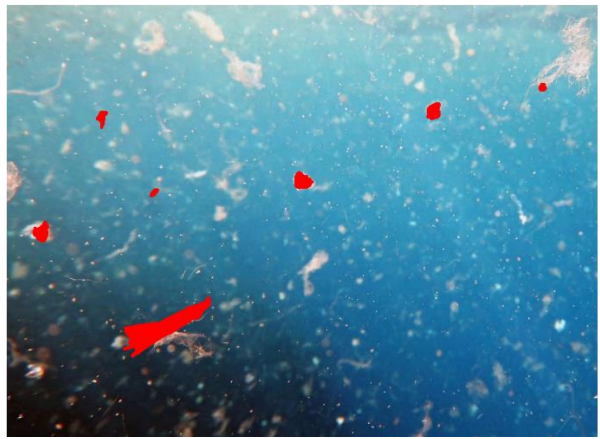

*Figure 10.2. Pipeline steps for object detection for debris using image segmentation methods that is then fed into a small neural network The image below the steps shows the algorithm identifying debris in a sample image by segmenting portions of it with red pixels.*

The images above show that the program algorithms had difficulty in detecting all the major elements of the image in terms of the debris. Results for both the bounding box and the segmentation results showed that it would be better for the images to be preprocessed and adjusted to a uniform definition for both shape and color before being fed into the neural network. Imaging sizes and pixilation yielded similar accuracy to the actual case for the bounding box algorithm (last tested ~64%). It was also observed that the addition of more layers improves performance but reaches a maximum beyond which no further changes are made. As a result, it was decided that to make the models more discriminatory by adding other types of waste types and then comparing those results against a classifier for microplastics or other plastic debris. This change in the training network, helped in making the network more accurate and adaptable by introducing different classes of wastes. In order to make the network more efficient in picking up plastic debris, some 'noise' was added in the form of additional classes and different waste types, including wood, metal, paper. Trash, cardboard and more. The images below show the results of rescaling which was performed on the images to make them more uniform for processing and classification.

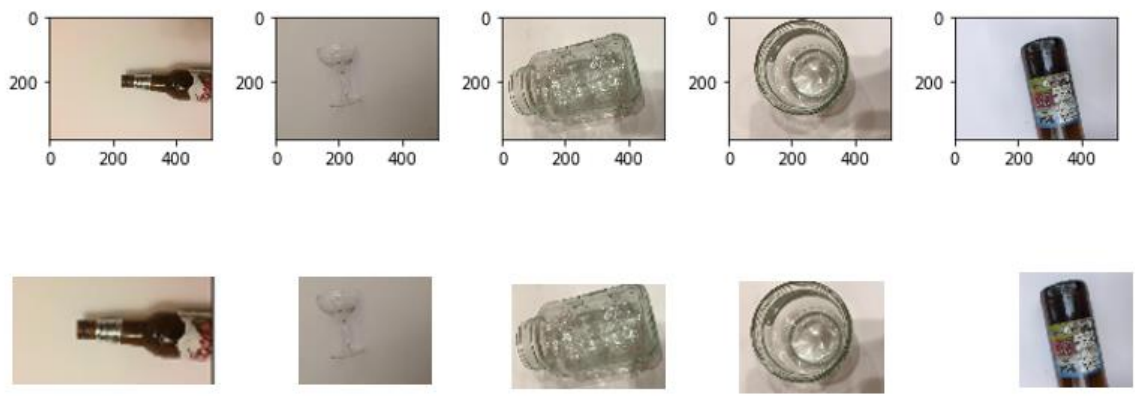

*Figure 10.3. Corrections in the depth and focal length of the images based on algorithms from pyimagesearch. The image series on the top is the original sampled dataset while the image series below shows the same images with the objects zoomed in and pixel polished for analysis.*

The image shown previously shows the results of a distance correction algorithm that factors in changes in focal length and focuses on the exact waste item to ensure that the image becomes more compatible for processing without losing its features. The package pyimagesearch was mainly used for this purpose. The result of the remaining images and their dimensional corrections can be seen in the figures below.

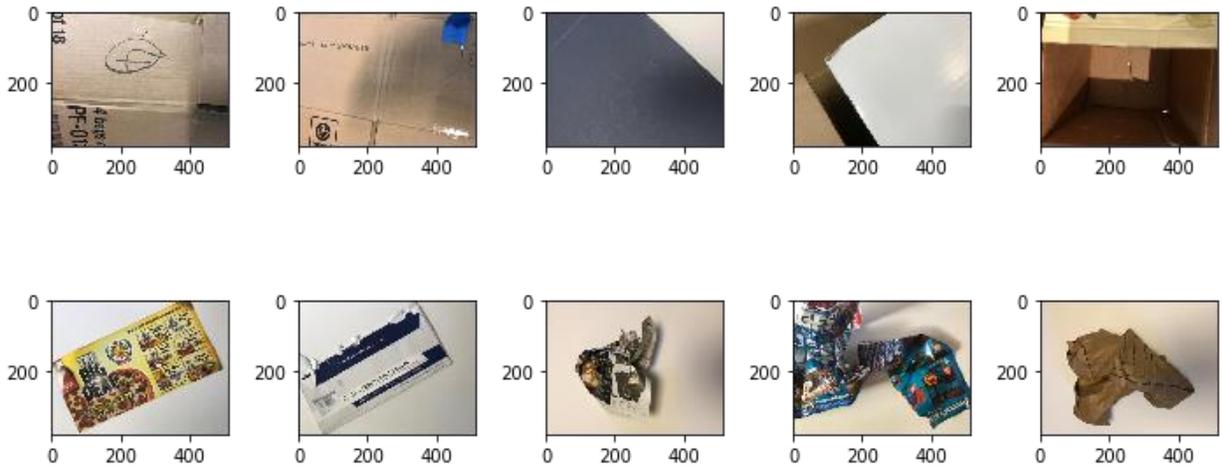

*Figure 10.4. The sampled image sets above show the corrected dimensions(width and height) that were used for making all the images uniform for preprocessing analysis.*

The images in the dataset were corrected for color, dimensions (height and width) as well as contrast and depth to ensure that all they are all uniform and prepared for analysis. Based on research conducted on other object detection algorithms in the past on similarly sized items, a standard size of 400x500 was selected for all the images. It was further decided to improve on the algorithmic ability to handle changes in the image depth, the bounding box shape was adapted to fit the waste item, instead of using a standard size which showed several complications in the past simulations, due to the limited spacing.

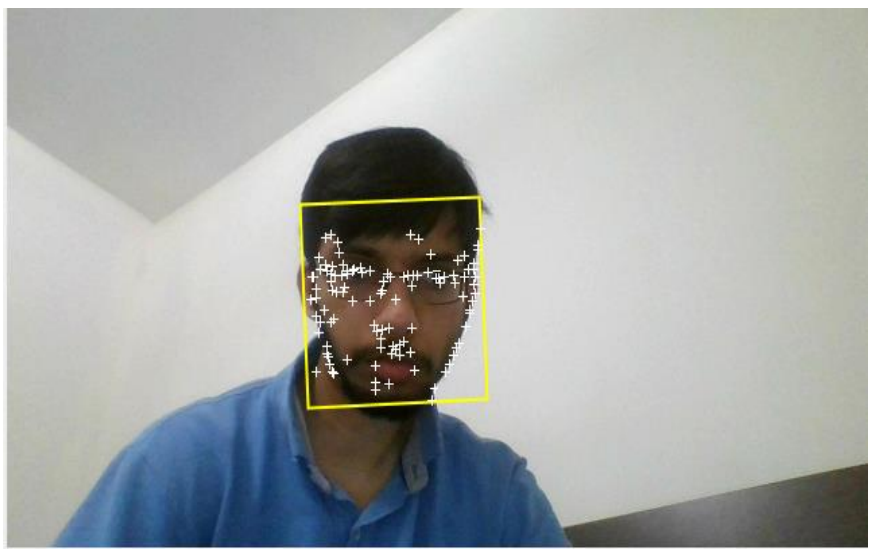

*Figure 10.5. Video sampled analysis of face tracking based on feature extracted points from facial shapes such as nose, eyes and chin. This image shows how detection algorithms use tracking points to study object features in frames and connecting them to the bounding boxes.*

Object tracking and feature extraction can also be achieved using a dynamic batching algorithm that allows the user to implement different batch sizes and hypermetric parameters. The figures above show the dynamic features that are extracted from facial images that track eye movements, nose shapes, facial symmetry and other perimetric data. Dynamic Batching feature allows the user to dynamically change batch size for inference calls within pre-set batch size limit. This feature might be useful when batch size is unknown beforehand and using extra large batch size is undesired or impossible due to resource limitations. For example, face detection with person age, gender, or mood recognition is a typical usage scenario. The user can activate Dynamic Batching by setting KEY_DYN_BATCH_ENABLED flag to YES in a configuration map that is passed to the plugin while loading a network. This configuration creates an ExecutableNetwork object that will allow setting batch size dynamically in all of its infer requests using SetBatch() method. The batch size that was set was then passed through the CNNNetwork where the dynamic object is used as a maximum batch size limit.

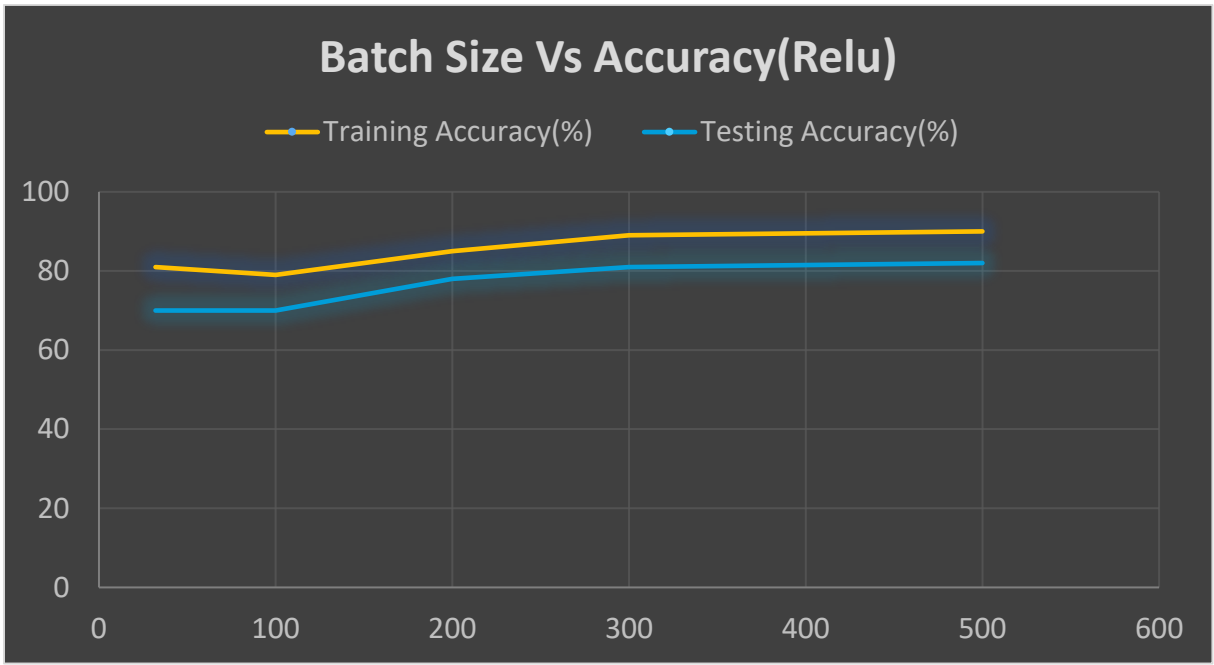

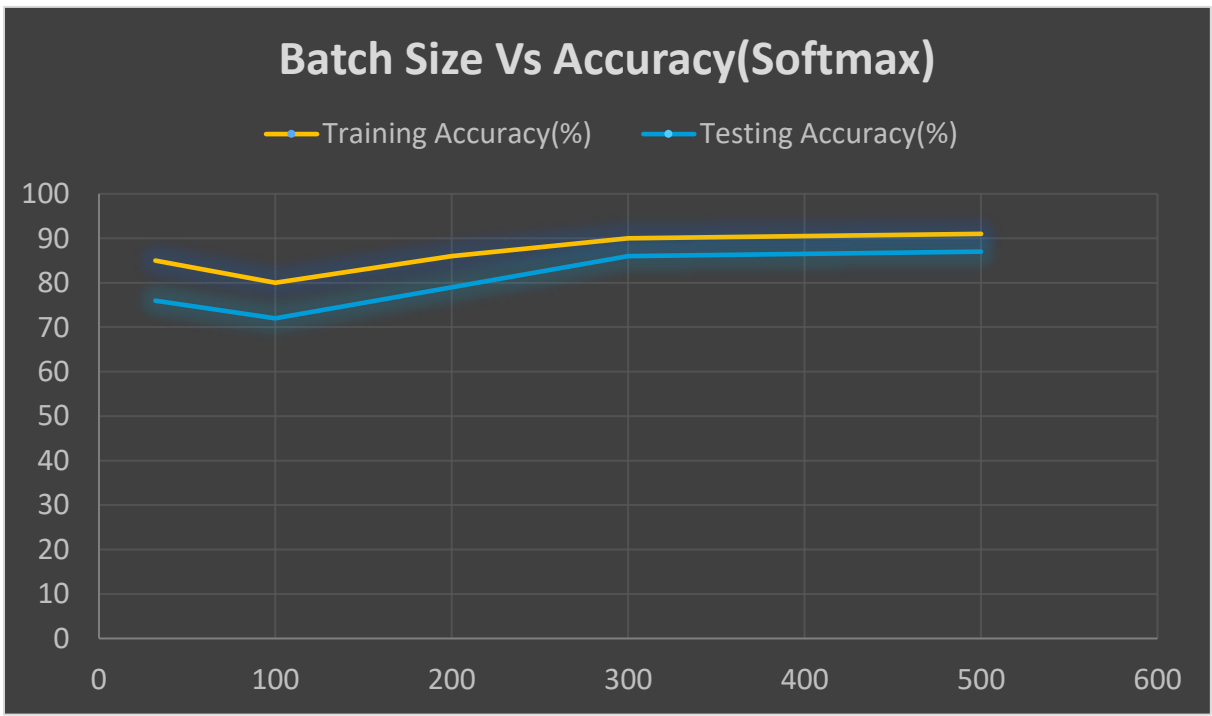

*Figure 10.6. Results for sampled image analysis tests using a typical CNN for both training and testing sets with ReLU and softmax activation functions. The figures above compare the accuracy results(Y-axis) and its relation to the dataset set(indicated on the X axis in terms of number of images in the set).*

In order to make the detection system more refined, it was concluded that conventional models like MXNet. COCO or ImageNet are not proficient in detecting waste types, prompting the need for a specialized CNN architecture consisting of multiple layers and nodes. Using a normal scheme for 70/30 split in training and testing datasets, the algorithms derived were tested against different batch sizes to assess how well the models perform. The graphs shown previously confirm that adding more images to the model makes it more powerful in extracting more features from the graphics. Tests were performed for both the ReLU and softmax activation functions. It can be seen that softmax yields better accuracies than conventional ReLu (rectified linear activation function) and that increasing batch sizes does increase model accuracies but reaches a threshold at a certain point. The size of the CNN was selected as 500x250x100 but was augmented to allow more datasets to be analyzed in future simulations. Further noise addition was implemented for improving on the accuracy.

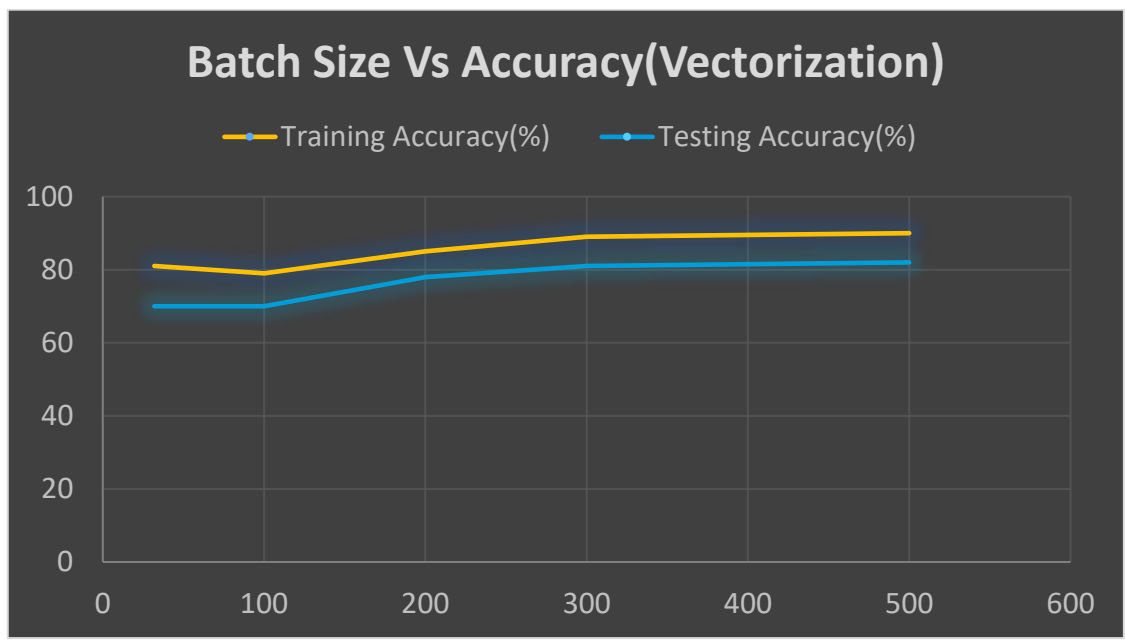

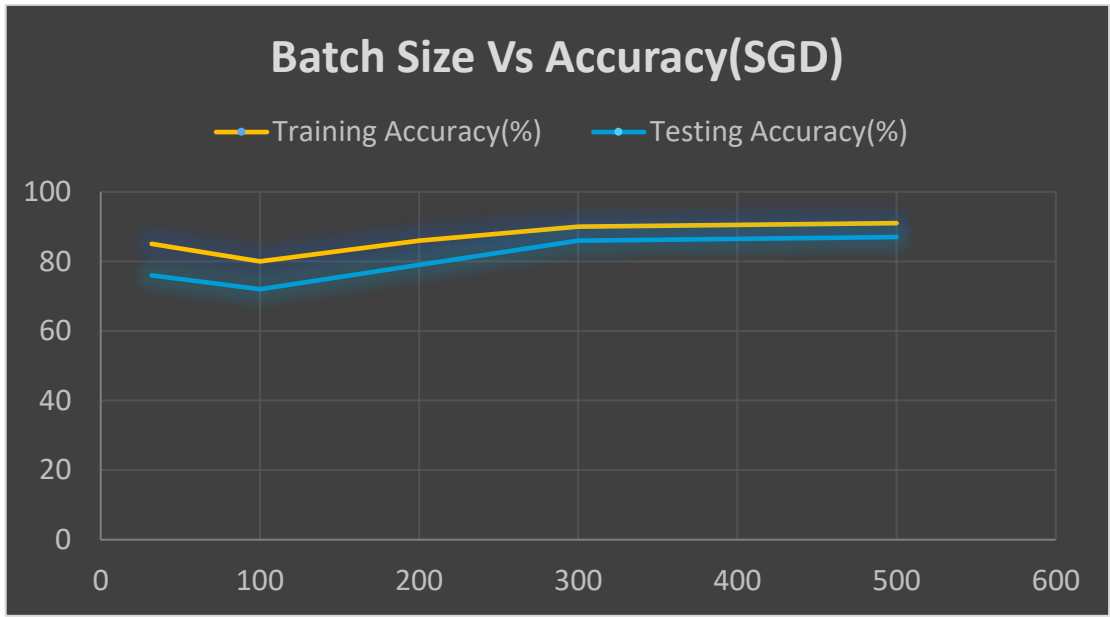

*Figure 10.7. Results for sampled image analysis tests using a typical CNN for both training and testing sets(with vectorization and SGD activation functions). The figures above compare the accuracy results(Y-axis) and its relation to the dataset set(indicated on the X axis in terms of number of images in the set).*

Further tests and simulations performed on vectorization and stochastic gradient descent algorithms showed that SGD works slightly better for plastic detection. Imaging effects like grayscaling and negative correction can only improve the accuracy to a certain extent for plastic detection(79% vs 82%). These improved architectures also helped in inserting a new batch of images to be processed with algorithms for corrections that target image dimensions including object size, focal length and image clarity. A standardization algorithm was further implemented to create a uniform depth of perception for images to have the same pixel density and clarity. The above step resulted in changes to existing images to become zoomed in or out to make them more uniform.

## 11. HYPERPARAMTERIC TUNING TESTS

OpenCV is one of the handful of platforms for accomplishing the results of this project. OpenCV uses the cv2.VideoCapture function that takes the input image or video file from the user as an object which is captured later using capture.open. Similar to a camera lens, the system can be used to confirm if it is open using capture.isOpened, and the user can read a frame from it with capture.read. This read function can actually return two items, a Boolean and the frame. If the Boolean is false, there's no further frames to read, such as if the video is over, so the user should break out of the loop

Once the user has used up the entire image, there are additional steps to undergo with OpenCV. First release the capture which will output the captured file or stream to the file space. Then shut down any windows using cv2.destroyAllWindows. For extra steps, add the cv2.waitKey within the loop, and break the loop if the user's desired key is pressed. We saw the cv2.VideoCapture function in the previous video. This function takes either a zero for webcam use, or the path to the input image or video file. That is just the first step, though. This "capture" object must then be opened with capture.open.

Then, the user can basically make a loop by checking if capture.isOpened, and the user can read a frame from it with capture.read. This read function can actually return two items, a Boolean and the frame. If the Boolean is false, there is no further frames to read, such as if the video is over, so the user should break out of the loop. Once there are no more frames left to capture, there's a couple of extra steps to end the process with OpenCV.

First, the user has to release the capture, which will allow OpenCV to release the captured file or stream. Second, the user needs to use cv2.destroyAllWindows. This will make sure any additional windows, such as those used to view output frames, are closed out. Additionally, the user may want to add a call to cv2.waitKey within the loop and break the loop if the desired key is pressed. For example, if the key pressed is 'Q', that is the Escape key on the keyboard - that way, can close the stream midway through with a single button. Otherwise, the user may get stuck with an open window that is a bit difficult to close on its own.

Video Capture - can read in a video or image and extract a frame from it for processing to resize and use it for future simulations with the same given resized frame. cvtColor can convert between color spaces. TensorFlow models are usually trained with RGB images, while OpenCV is going to load frames as BGR. There was a technique with the Model Optimizer that would build the TensorFlow model to appropriately handle BGR. If the user does not add that additional argument, this function can be used instead to convert each image to RGB, but that is going to add a little extra processing time. rectangle - useful for drawing bounding boxes onto an output image inwrite - useful for saving down a given image.

Building important insights and details from existing models is necessary in order to derive other useful metrics from detection schemes. The code sample below is an example of one such algorithm observed from literature for object detection:-

```python
def assess_scene(result, counter, incident_flag):
    '''
    Based on the determined situation, potentially send
    a message to the pets to break it up.
    '''
    if result[0][1] == 1 and not incident_flag:
        timestamp = counter / 30
        print("Log: Incident at {:.2f} seconds.".format(timestamp))
        print("Break it up!")
        incident_flag = True
    elif result[0][1] != 1:
        incident_flag = False
    return incident_flag
```

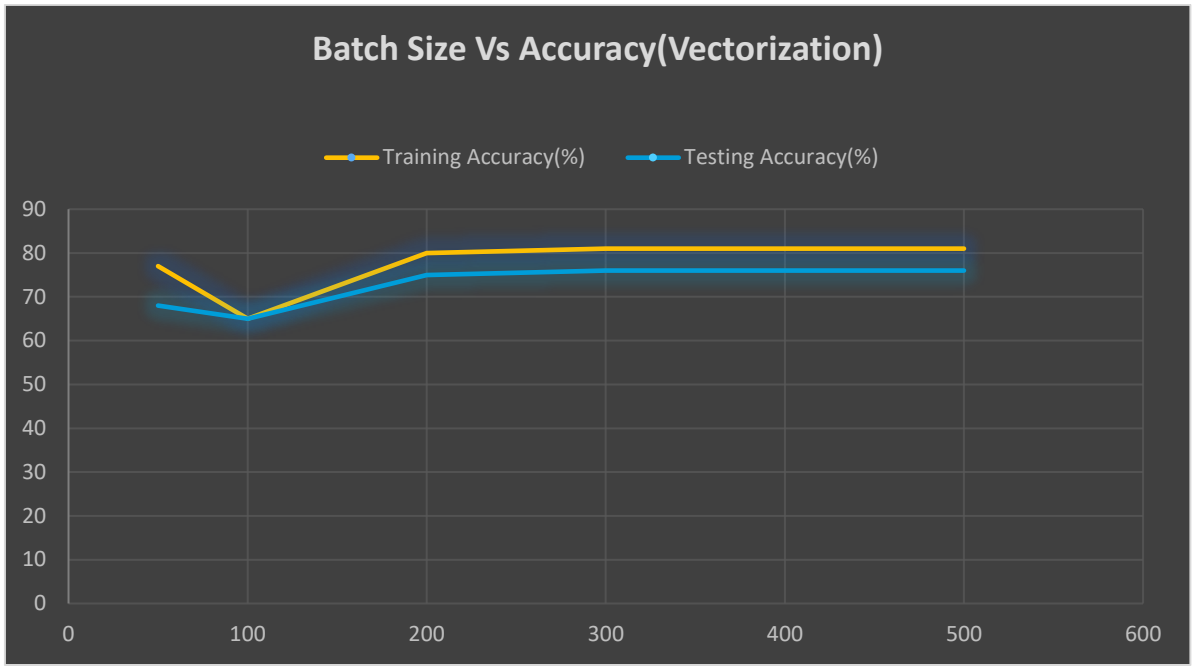

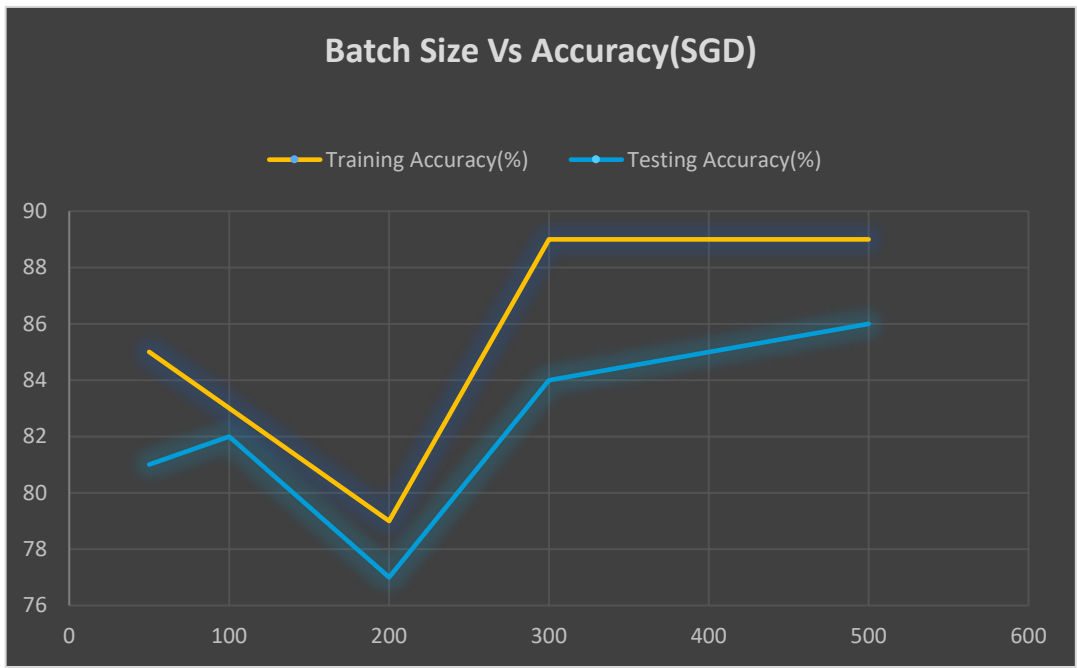

*Figure 11.1. Results for sampled image analysis tests using a typical CNN for both training and testing sets with colour corrected and density corrected alterations(with vectorization and SGD activation functions). The figures show that not all preprocessing steps are crucial for analysis and some can affect the network's accuracy statistics.*

The figurative results from implementing these hypermetric tuned results show that while the SGD algorithm has difficulty in maintaining accuracy, it picks up at larger batch sizes and consistently maintains it for both testing and training datasets. The vectorization algorithm is better in incorporating these parameters but does not fare too well in adapting these parameters with all kinds of testing datasets. Despite the black box nature of the datasets, we can include some of the extracted features against the classified images as inputs and check how important they are in detection. The density and lighting correction changes seen in the figures in the next page illustrate these changes. The same changes were inserted into the images after applying the preprocessing steps to allow the algorithm to create distinction between the object(waste item) and the surrounding to a clearer degree. This is also highly dependent on the size of the kernel.

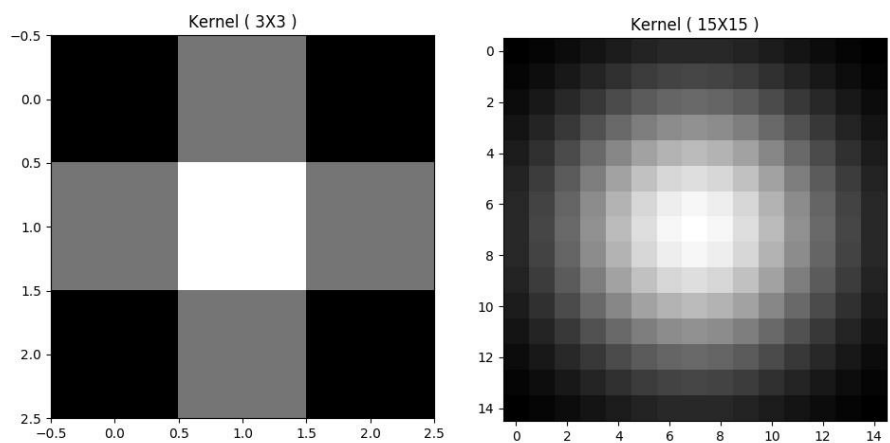

*Figure 11.2. Example of HOG variant corrections applied to images. Notice the change in pixelated points , which help the network to better identify objects with unique outlines when compared to the rest of the image, even without colour. (Source:-Blog)*

The results and tests show that color plays a huge role in determining the type of waste and may have an influence on the role of pixel breakdown when converting bands across the neural network. Accuracies for detection with a dataset of 200 images of different waste types produced weak results. Excessive use of filters is not recommended in particular cases when the images may become indistinguishable beyond comprehension. Gaussian filters that highlight the outline of the waste objects are far more useful than filters that hamper contrast, focus or add blur to the image. In terms of single shot detection algorithms and bounding box algorithms, the latter are far more useful for tests due to the convenience of coding unique detection schemes and pre-processing changes in the pipeline before the images are analysed. A majority of these changes were made possible from the packages available in OpenCV but were coded using Python. Some of the changes of input depths, filters and colour corrections are displayed in the figures below.

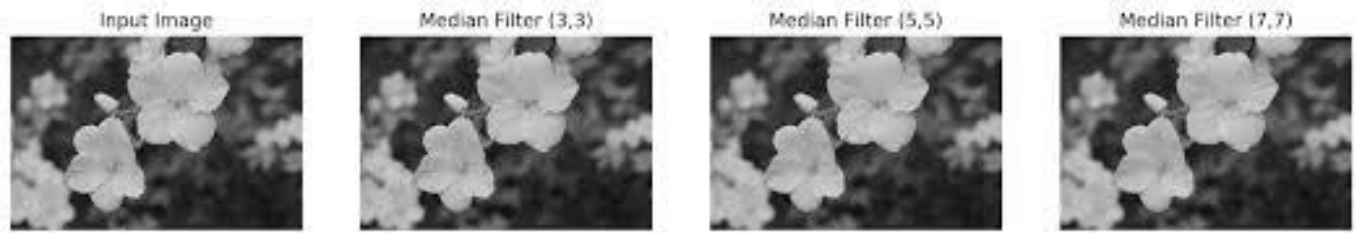

*Figure 11.3. Sampled result of median filters applied to an image to further separate the perimeter of the object from the surroundings. The median filter can be adjusted based on user requirements but tend to contribute towards lower accuracies if not applied properly. (Source:- Blog)*

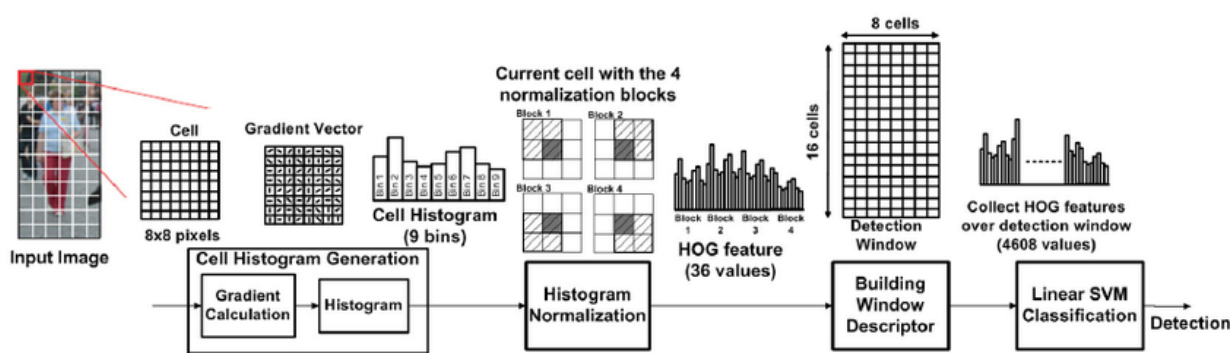

*Figure 11.4. Steps for object detection in images using HOG vectors and Linear SVM Classification, similar to a segmentation algorithm. [14]*

# 12. CONVLUTIONAL NEURAL NETWORK SIMULATIONS

After having the simulations for convolutional neural networks setup for image analysis and object detection, a portion of the code was set out for testing out a more commonly used application(people counting in frames) and also for waste detection through the Autocaffe models. The two figures below show a potential application of using a counter UI that can be used for tracking waste debris instead of people. The accuracy and frames analyzed per second are highly determined by the hardware and computing powers used for this analysis which can be seen through the differences in the inference times of the videos. The average duration of detection is highly variable on the hardware as well and does not remain consistent for all kinds of hardware, software builds. This further illustrates the importance of performing tests on different hardware component types, if the algorithms and models are to be executed in more realistic environments. This is even more significant if this execution requires replication of the algorithms in multiple testing sites, constrained by spacing and purchasing costs.

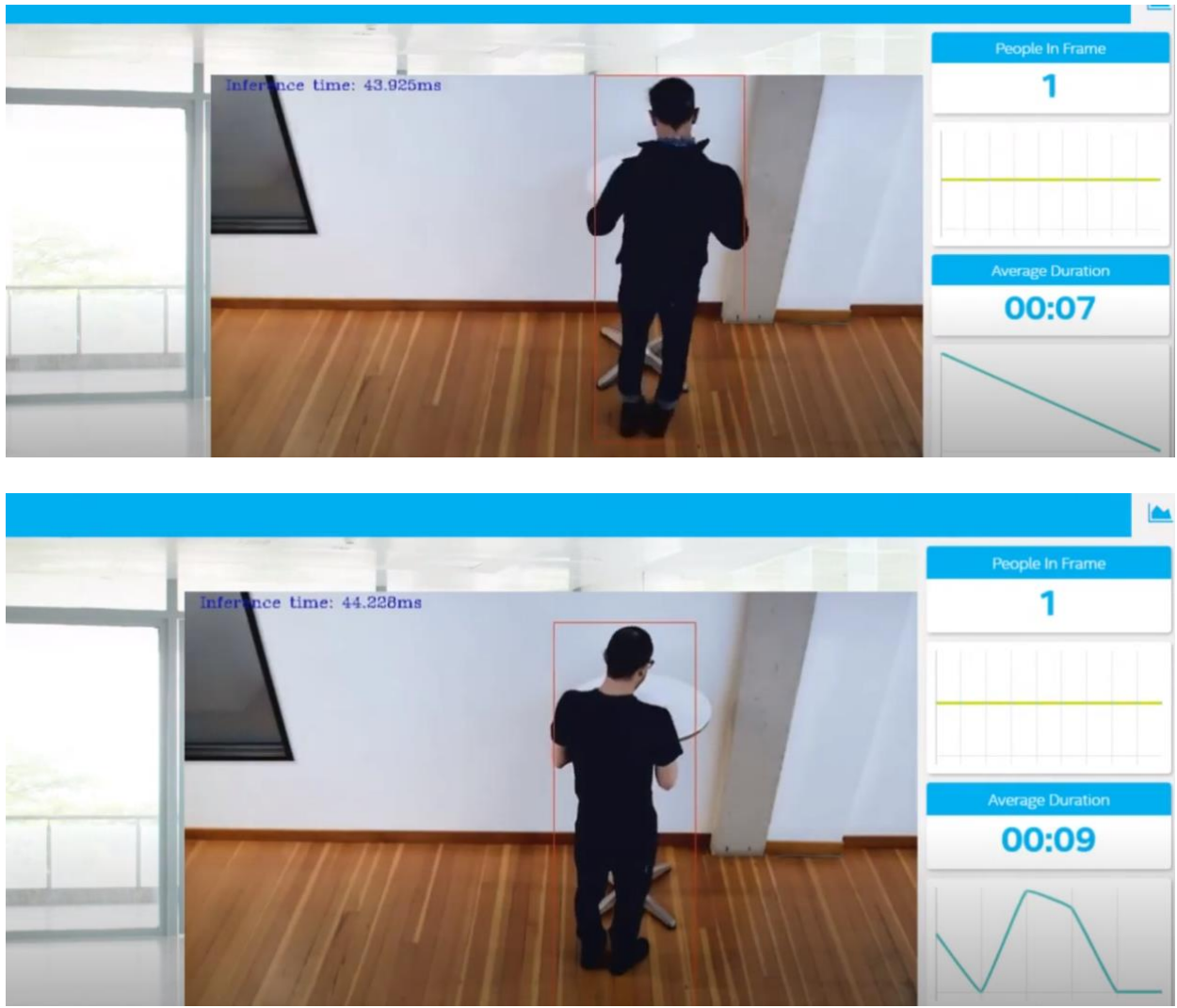

*Figure 12.1. Simple people counting application built using Open Vino with a useful user interactive output. This can be a steppingstone to creating a tracking and user interactive platform for detecting plastic debris. The form also shows the inference time, bounded box, number of detected entities(people) and the average duration of detection.*

The same CNN architecture discussed in the previous sections was amplified with better nodes and coefficients as the simulations were executed. The figures below are the results of one the simulation runs that uses the entire dataset, indicating that the algorithms are quite powerful in reducing the losses as more and more images are added to the pipeline. And this is same for both the training and the testing datasets.

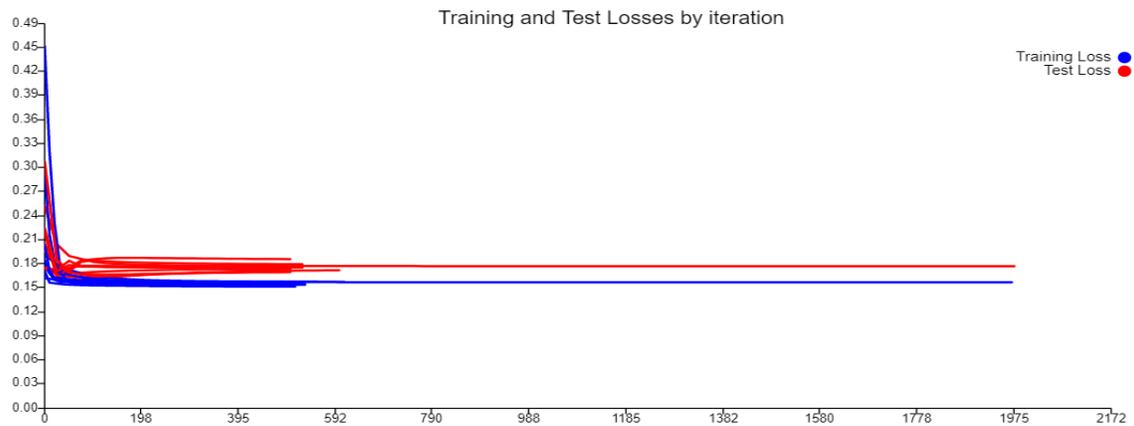

*Figure 12.2. Improving performance in the use of Adam and SDG solver functions for object detections with increasing samples sizes, related to both testing and training. Readers can see how the losses decrease while holding out a maximum value at lower samples sizes.*

The additional information for the simulation runs and the comparisons can be better seen from the figure below where the mean and median losses are reduced with different sizes, iterations, solver-types(activation types) and neural network sizes. To summarize, these models are slightly vulnerable to overfitting and have to be better validated with other datasets. However, repeated runs with different shuffled datasets indicate that the overfitting may be avoided altogether by simply changing the activation functions.

```
Test losses statistics
Mean   : 0.177993   Max  : 0.188233
Median : 0.177521   Min  : 0.171767
SD     : 0.004871   Range: 0.016466
(Mean-Min)/SD: 1.278311
```

| Config# | Test Loss | input-size | iters | solver-type | nn-size |
|---|---|---|---|---|---|
| 1 | 0.177537 | 2 | 2000 | Adam | 8 |
| 2 | 0.181777 | 2 | 2000 | Adam | 16 |
| 3 | 0.177506 | 2 | 2000 | Adam | 16 |
| 4 | 0.188233 | 2 | 2000 | Adam | 16 |
| 5 | 0.180881 | 2 | 2000 | Adam | 32 |
| 6 | 0.171767 | 2 | 2000 | SGD | 8 |
| 7 | 0.174148 | 2 | 2000 | SGD | 16 |
| 8 | 0.172995 | 2 | 2000 | SGD | 16 |
| 9 | 0.175724 | 2 | 2000 | SGD | 16 |
| 10 | 0.179366 | 2 | 2000 | SGD | 32 |

*Figure 12.3. Comparison model results for different simulations based on the activation function, input sizes, iteration numbers, solver types and neural network sizes. It can be seen that after achieving a maximum, the system reaches a threshold beyond which the accuracy or test loss cannot be changed. The SGD solver appears to have higher losses making it appear as better, but it only indicates the possibility of an overfitting case.*

Individually, the training and testing predictions are visualized from the figures below where the lines represent how far the system deviates in detecting the actual image type. These figures relate to the Adam solver algorithm which proved to better at avoiding overfitting against the training and testing portions of the dataset, while also adapting well to different types of waste types with minimal interferences in network strength. The lagged correlations too take a dive after reaching a maximum value at a certain batch size of images, for both the training and testing portions of the datasets inserted into the pipeline.

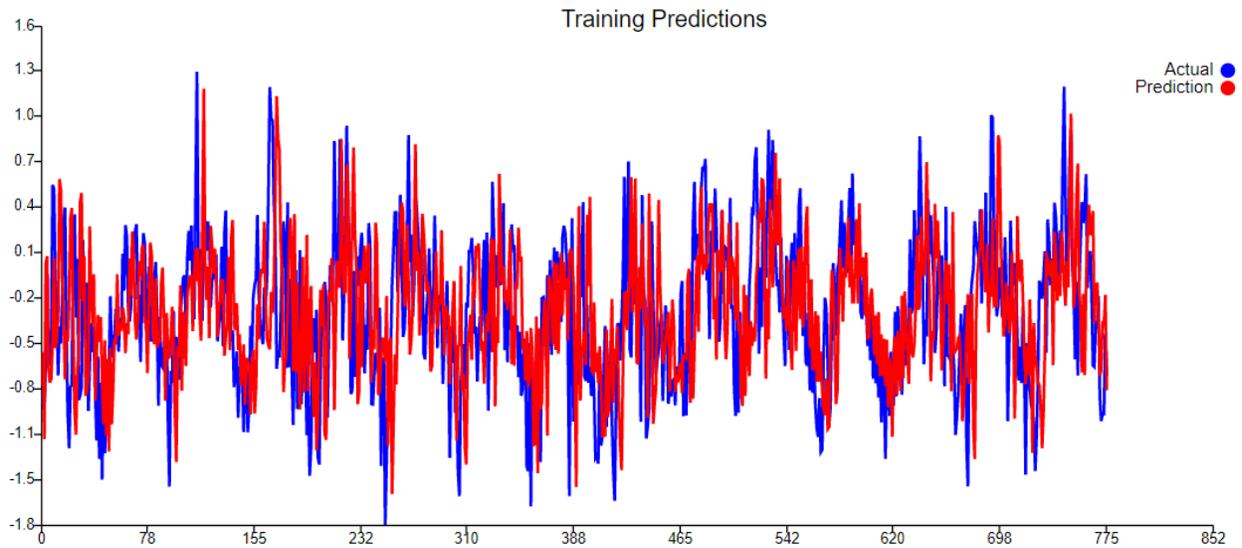

*Figure 12.4. Actual and predicted accuracy results with X-axis representing the image number and the Y-axis representing the deviation from the actual expected classification result(a zero or median value representing a fully accurate result with no bias or variance).*

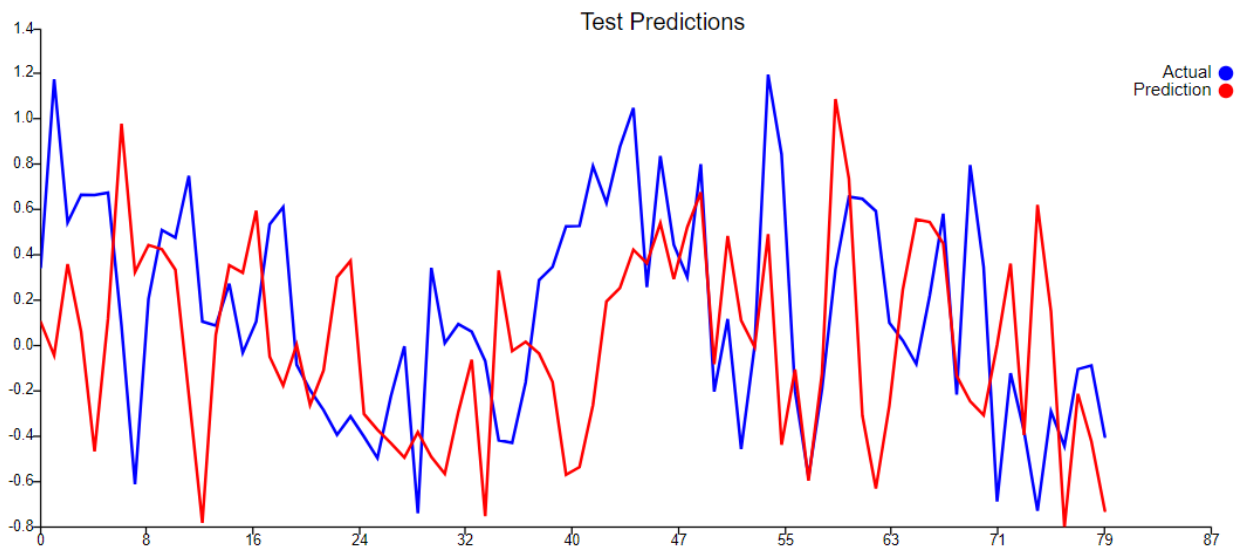

*Figure 12.5. Test prediction deviations from actual labels to predicted labels for both solver types. This figure illustrates how the network can become vulnerable to sudden changes in the image density, colour and pre-processed alterations. The X-axis represents the image number and the Y-axis represents the deviation from the actual expected classification result(a zero or*

*median value representing a fully accurate result with no bias or variance)*

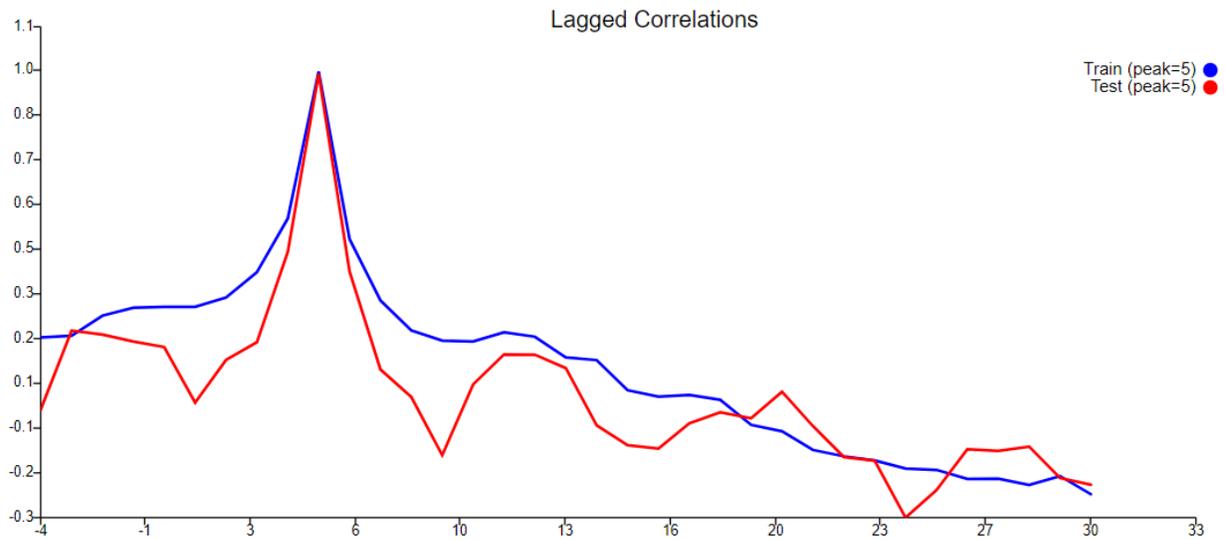

*Figure 12.6. The lagged correlations results between the training and testing sets. The figure shows that the network peaks around the fifth cross validation to produce reach a threshold value. After this value, the trends drop to a value below zero with more images in the batch. This can be made better with an extra layer or by combing ensemble activation functions.*

The images below contrast the results obtained from the simulation runs of the SGD algorithm against the Adam algorithm activation function. While the lagged correlations follow a typical bell curve, the most notable difference is the total overlap between the training and testing predictions which may indicate that the algorithm is well balanced. However, this only points towards how the SGD can become subjective to overfitting, even for completely balanced datasets. The verdict thus is that the network should contain a combination of both if possible, with a greater focus on the Adam activation functions. The SGD is still highly powerful in learning quickly, even when the dataset has a high number of ground truths to study and distinguish. Overfitting as a whole can be prevented by changing the distribution of the datasets used for training and testing and allowing the system to learn incrementally with a rising training to testing ratio. However, the model will reach a threshold, as previously observed but this shouldn't deter the user from improving upon it.

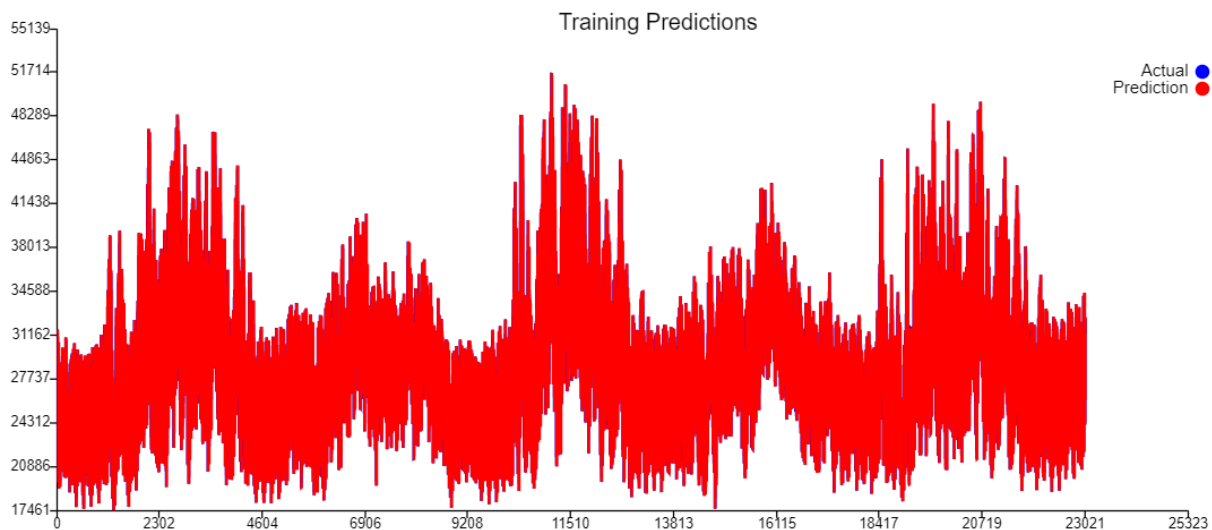

*Figure 12.7. Actual and predicted accuracy results with X-axis representing the image number and the Y-axis representing the deviation from the actual expected classification result(a zero or median value representing a fully accurate result with no bias or variance). These results are reflective of the Adam function's tendency to overfit nearly on all the images in the batch.*

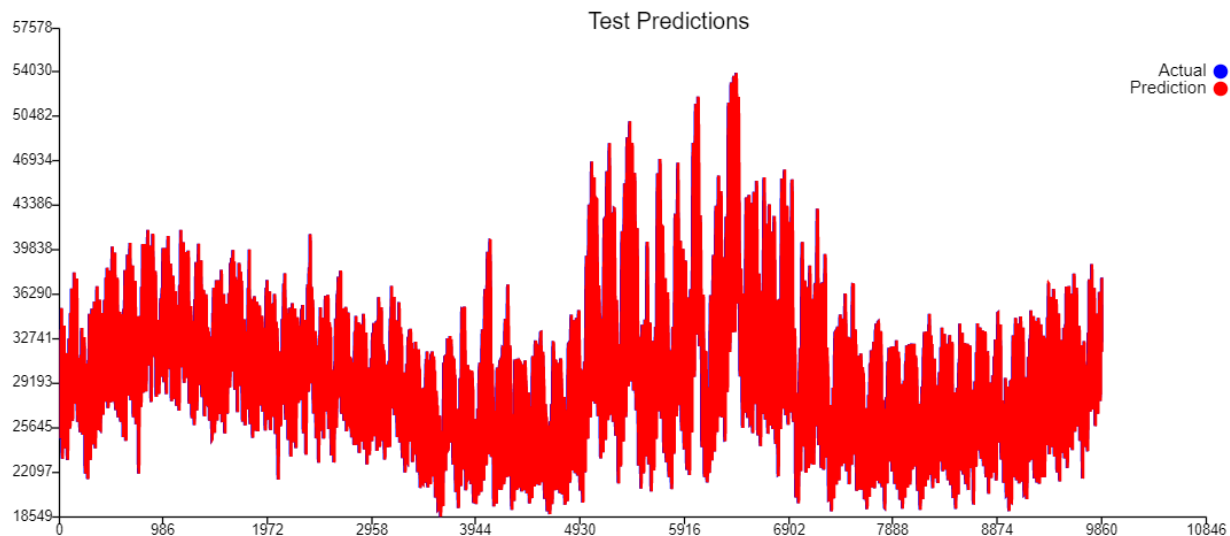

*Figure 12.8. Actual and predicted accuracy results with X-axis representing the image number and the Y-axis representing the deviation from the actual expected classification result. This figure does prove that while the Adam function may be highly accurate in predicting image types even in the testing set, it can be proven only after testing the model with more image batches.*

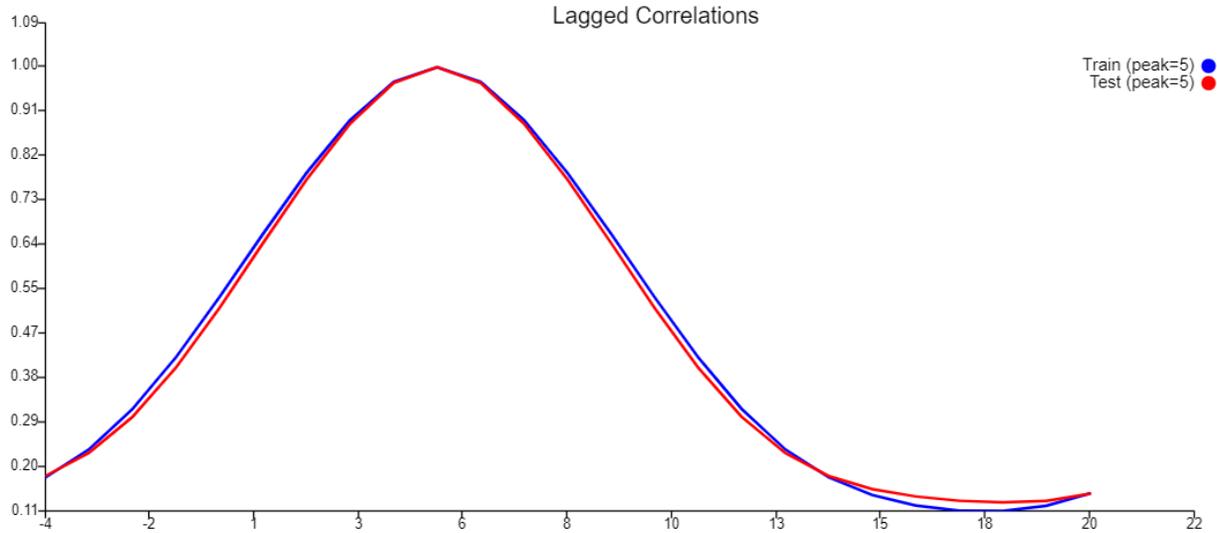

*Figure 12.9. The lagged correlations results between the training and testing sets. The figure shows that the network peaks around the fifth cross validation to produce reach a threshold value. After this value, the trends drop to a value below zero with more images in the batch. The figure shows that the Adam function is more streamlined in lags when compared to the softmax figure where training and testing correlations increase and decrease irregularly.*

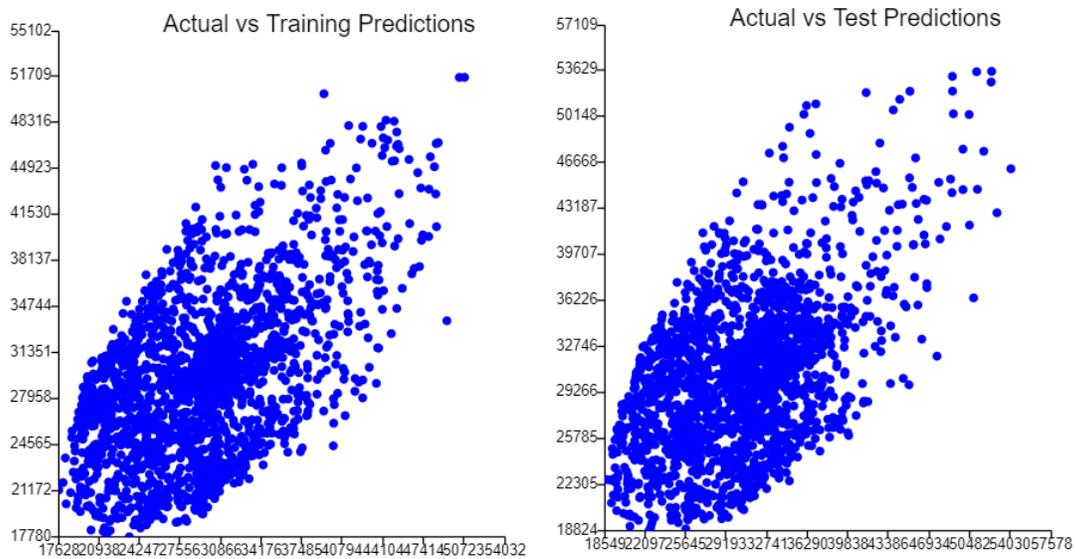

*Figure 12.10. Frequency distribution plot for image points against actual and test predictions. The figure on the left is obtained from the softmax tests while the image on the right is obtained from the Adam activation tests. Notice how the distribution is quite similar, despite the differences in correlations seen in the previous figures.*

# 13. FUTURE IMPLICATIONS-HARDWARE AND LIMITATIONS

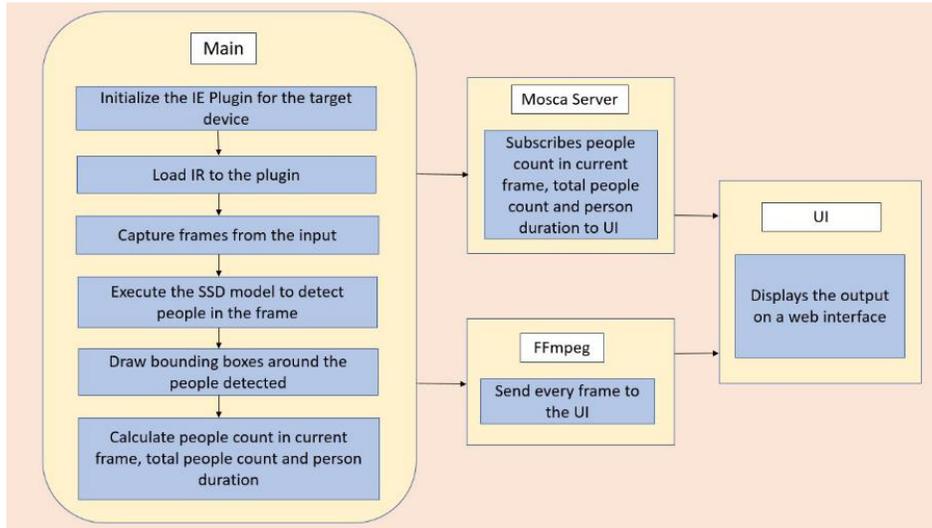

*Figure 13.1. The steps involved in setting up a server through Open Vino for counting algorithms, linking it with a loaded model, processing a sample video frame and displaying it through a user interface.*

This section of the project aims at discussing some potential solutions for hardware implementations that would host the algorithms and track plastic debris. The solution of using computer vision for object detection and to an extension for segregating plastic wastes, warrants the need for exploring options for machines that can achieve this at larger levels. In order to test out the best suited hardware technologies that could be used for this purpose, three main case studies were considered and executed on the OpenVino platform. The hardware components tested include a simple CPU, FPGA, GPU and the less common VPU. In more realistic formats, it is necessary to account for factors such as model loading times, inference and the total frames analyzed per second. These factors are important because of their relation to the maintenance, cost and setup associated with each of the hardware, allowing researchers to better select the most optimal hardware for testing out the algorithms. These factors also help to visualize secondary factors such as heat losses and network pack losses. It is important to add that these models were more focused on object detection as a whole and not specifically plastic waste. However, due to the similar nature of the materials and the sizes, the comparison results can be related to plastic debris as well.

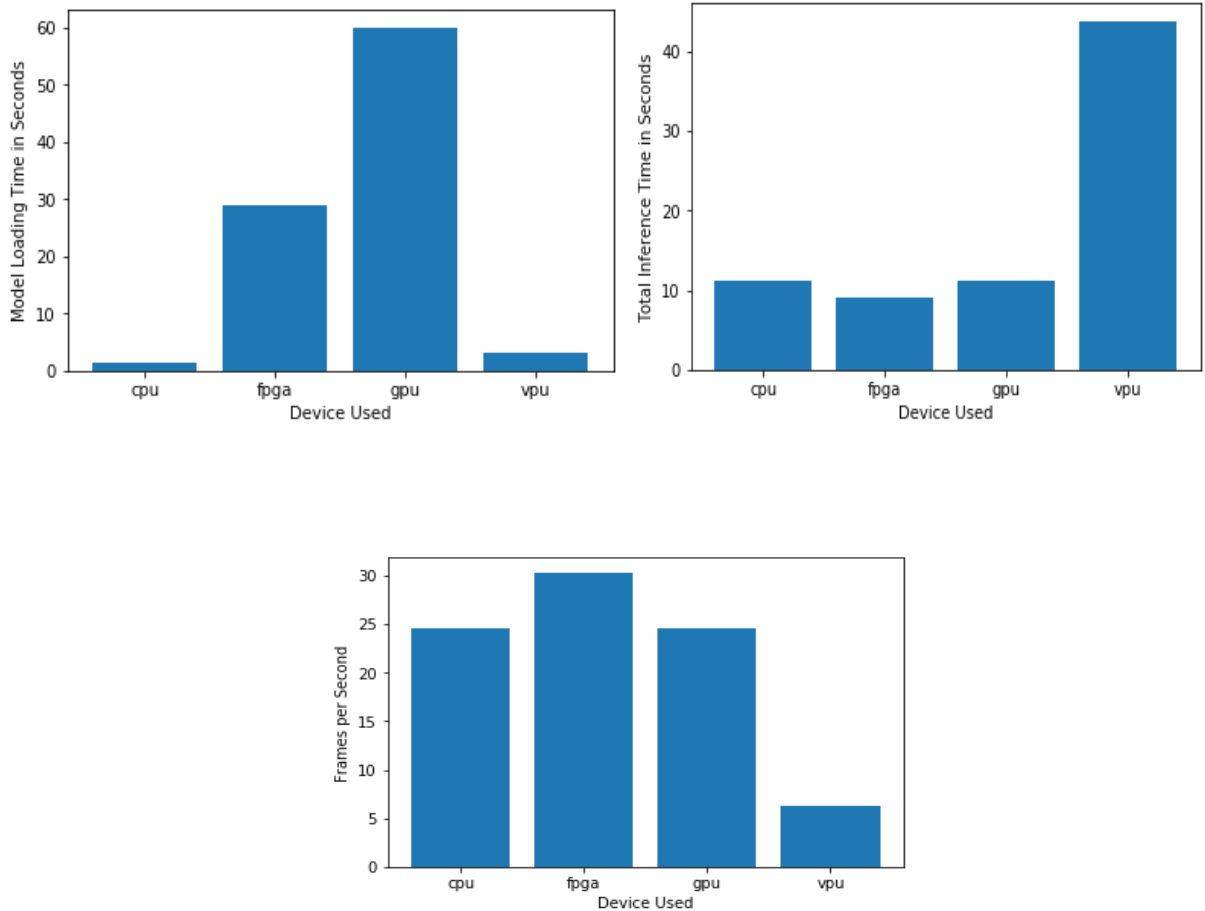

*Figure 13.2. Model loading time, total inference time and analysed frames per second for the manufacturing case study compared against a CPU, FPGA, GPU and GPU. All results were obtained using the Open Vino platform from Intel on an online server.*

The results from the first case study was conducted on a manufacturing video where objects of different types had to be detected. The video feed was setup using a normal handheld video camera, placed at a height. The results show that FPGA is perhaps the best default choice for this task, due to the fast inference times along with the small space. They can be reprogrammed quite easily if the user wants to change configuration settings. Long durability and flexibility are also major plus merits. The FPGA is also proficient if the project objective is to perform video processing multiple times per second, with a modest consumption rate(100 mA to 60 A at 0.9 to 1.V giving an average wattage of 50-60W).

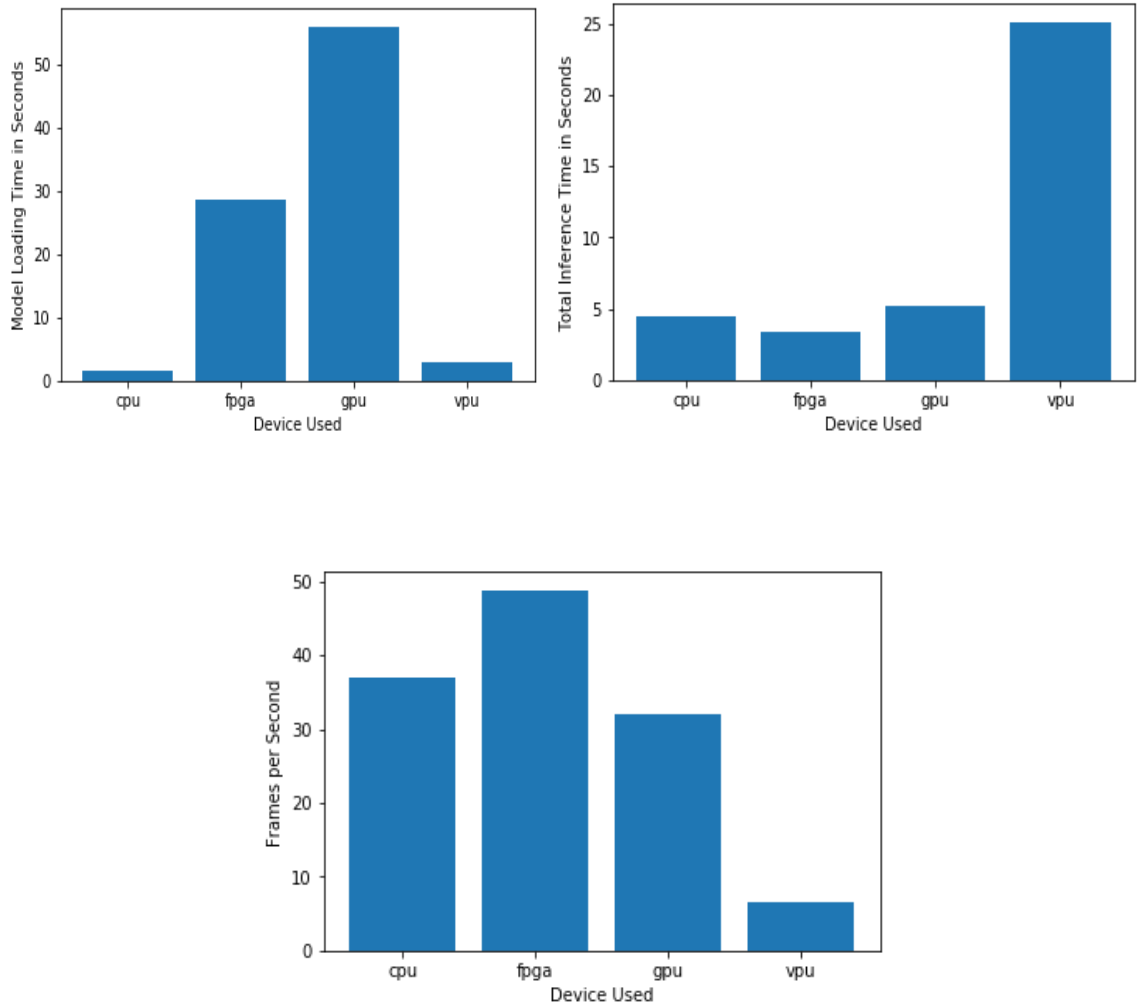

*Figure 13.3. Model loading time, total inference time and analysed frames per second for the retailing case study compared against a CPU, FPGA, GPU and GPU. All results were obtained using the Open Vino platform from Intel on an online server.*

The second case as seen in the graphs above is derived from a retailing case study that used a CCTV to track for both people and objects. This case study shows that if the user intends to run the detection algorithms on a budget, constrained by costs, the pre-installed intel i7 CPU is the best choice given its performance for detection and inference. Not all CPUs would contain an in built IGPU, which would further improve the performance of the system if the number of people increase on certain time periods. However, the final verdict remains that the CPU is the most optimal choice among the choice due to its limited consumption and little need for additional hardware. Giving away a typical wattage of 65-85 W, multiple units can be set up at different counters in the environment where the detection is to occur. With a usual 20 W thermal design power(TDP) and feasible wattage, the CPU nets feasible costs making it the best choice.

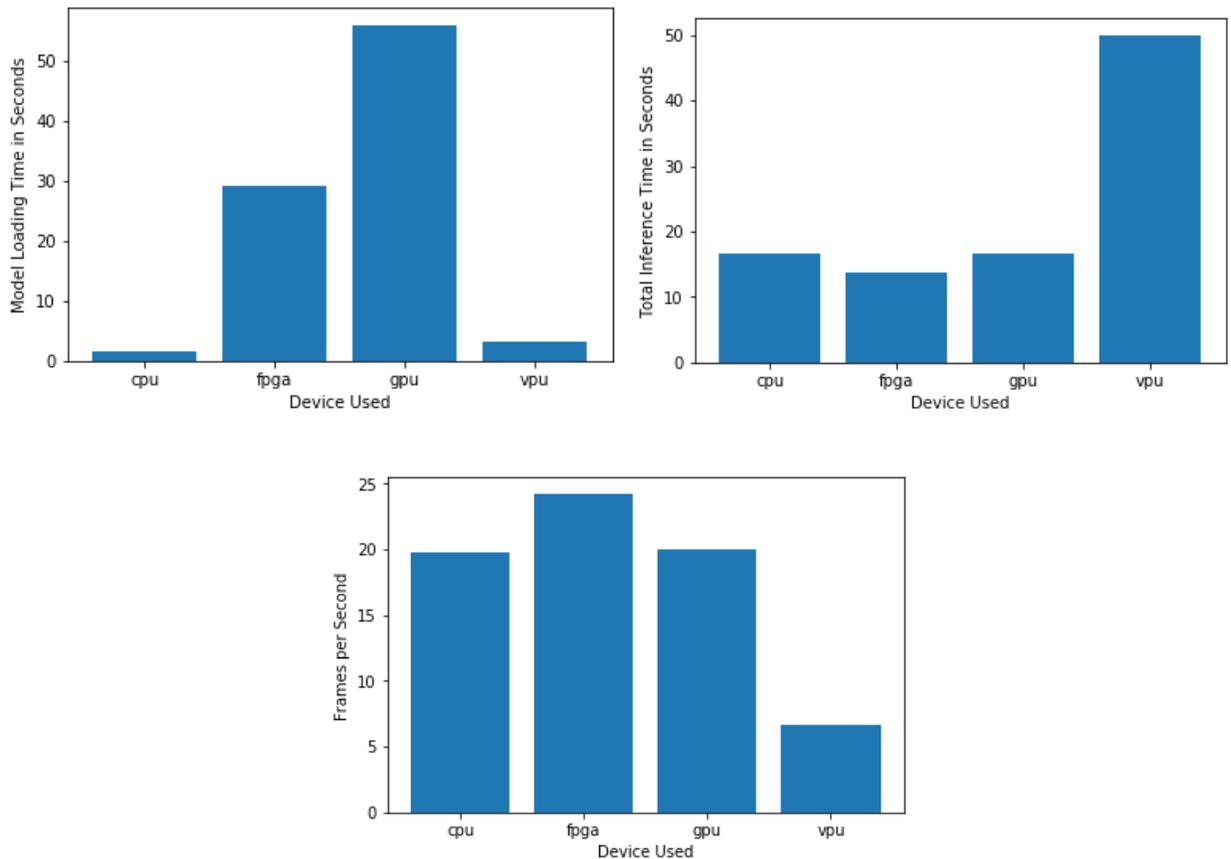

*Figure 13.4. Model loading time, total inference time and analysed frames per second for the industrial case study compared against a CPU, FPGA, GPU and GPU. All results were obtained using the Open Vino platform from Intel on an online server.*

The third case study concerns objects being carried onto a pulley moving from one end of the municipal unit to another stream. The video was sampled from an industrial video camera placed atop the pulley that has limited capabilities and vision depth measurements. Performance wise the FPGA holds up well for taking care of any detection and inference needs but the limitations of the budget and the space make this unfeasible, if the user is limited by both. The Myriad X and the NCS2 with a VPU are still a choice to be considered. However, if the project has a budgeting constraint, 3 to 4 VPUs can be bought and connected, costing close to $100 individually. The VPU in this case gives a near 50 second inference with the second least model loading time(~3 seconds) and has a decent FPS(7 FPS). Combined with the average wattage of a VPU unit and the fixed inference which cannot be increased, especially during peak hours, the VPU is the best choice for hardware recommendations.

The models used for these comparisons including the averaged effects of using both the Adam and SGD frameworks for optimization. It can be seen that Adam and SGD vary hugely in their training and testing losses. Adam models are faster at convergence but have too much lag in processing hyperparameters. Difference networks that measure difference between actual and expected output are preferable when determining colour differences. Non-persistent models like Naïve Bayes don't account for recent data and thus aren't too feasible in the context of detecting plastic debris. To conclude from the different case studies and hyperparameters that are studied in this section, FPGA and typical CPU configurations are still suitable for on time tracking at facilities within a budget, while VPU and GPU can be more costly.

## 14. CONCLUSIONS AND FUTURE WORK

From the simulation runs discussed in section 10, section 11 and section 12, we can infer the following results:-

1. Multiclassification models are more capable in adapting to new data as evidenced by the tests between the softmax and vectorization activation functions. While the testing and training accuracies take a hit with increased batch sizes, the trend reverses itself(as seen in Figure 11.1).
2. The Adam and SGD models are more compatible with identifying and tracking smaller objects. A particular pitfall, however, is a natural fall in accuracy if the image frame becomes too populated with non waste items. The solution to this is to continue tuning the algorithm(as seen in the SGD case with hypermetric tuning) with new datasets.
3. Non-ensemble methods may have been the first approach but are vulnerable to erroneous detection, especially for moving frames. Tests conducted on the OpenVino platform done using multiple layers of functions like kNN, detected only some waste items. However, numerous steps of image preprocessing were needed to fulfill this, making it unfeasible.

4. OpenVino tests show that edge-based applications can be created to track plastics at remote sites, as per the case studies done on various hardwares. Depending on the costs and maintenance involved, the project has given context for the best options for setting up hardware and the relevant algorithms.
5. The models developed have achieved a sizeable accuracy in detecting wastes using a normal set of preprocessing steps, rather than focusing on heavy image manipulation as seen in the preliminary tests in RStudio(Section 10). The models were built for waste detection and were later implemented on microplastics, yielding fairly consistent accuracies.
6. Softmax and ReLU have similar performances but lack the speed and inference capabilities to detect and draw bounding boxes around the object, especially in video feeds. This result is pivotal because it helped narrow the best model choices.
7. Simulation tests show that training and testing losses(for both Adam and SGD) drop rapidly and reach a reasonable threshold. Neural network size plays a role in minimizing this loss but is not the only factor that can be adjusted to improve accuracies.
8. Tests conducted on a batch of images with distance and focal corrections on objects indicates that there is ~1 to ~2% change in accuracy. These corrections do not improve adaptability to testing sets.
9. There are only a finite number of features that can be attached to objects to track them. Small number of features(5 to 10) can drop the accuracy if the objects are mobile. Large number of features can lead to an underfitting case(observed when using SGD in tracking floating plastic pieces). The optimal number of feature points is observed as 67 to 75.
10. An extremely large CNN is not necessary to achieve good accuracy in detection if the right kind of preprocessing corrections are implemented. Median filter changes, negative correction, Gaussian blurs, image resizing are the preferred methods for microplastic detection from the current simulations.
11. In terms of segmentation, the models perform with minimal segmentation preprocessing steps. Algorithms like Adam and ReLu perform better with segmentation in detecting objects but are highly overfit. SGD is still favoured for all cases, above other types discussed in this report.

Plastic debris is a highly underrepresented problem that requires greater attention due to its negative impacts on ecosystems and the quality of life for both humans and animals alike. Employing manual operations for tracking and detecting wastes is a step in the right direction but can be improved by bringing technologies such as open vision to the mix. With a growing interest in the same problems seen across the board by NGOs and startups, it is necessary to cultivate a repository of models and datasets that are catered towards removal and tracking waste. The models discussed in this paper, including the methods and approaches taken to arrive to them are aimed to bridge this gap of knowledge and computing resources, in order to facilitate cleanup projects and study the relations between human activities to plastic waste blooms in water bodies. With research still ongoing to make the models more accurate with varying architectures and algorithms(using different activation functions), the findings and conclusions of this project should shed some light as to how solutions can be built and what are the logistical merits of choosing certain hardwares against others. It is important to consider that the analyses were performed using a mix of datasets for both object wastes and counter tracking(that involved normal items and people). The results from the simulations confirm that SDG and Adam are better in building algorithms for waste detection and this can be extended to videos and other media formats(which are being tested currently). Further improvements have been made through feature extraction methods by using the augmented models against smaller datasets of a single type of waste. These models were tested for their model accuracy and visualized through a Gaussian XGboosting algorithm which can be seen from the figures below. The first figure shows the false positives, false negatives, true positives and true negatives from the Adam algorithm while the second figure relates to the SGD algorithm. The improved SGD algorithm still manages to be better in performance despite the previous vulnerabilities of overfitting.

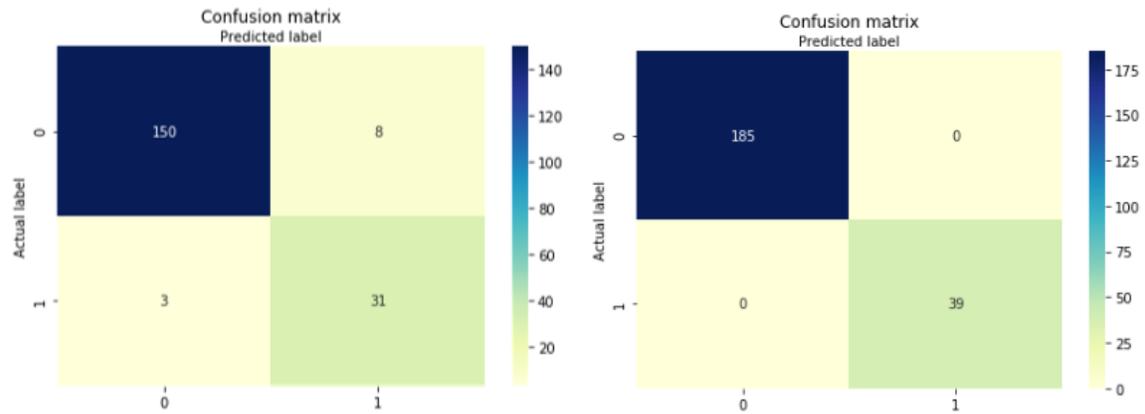

*Figure 14.1. Confusion matrix for images correctly and incorrectly detected using Adam and SGD activation function. The blue square in the top left are the images that were correctly classified and the square at the bottom right indicates the dummy images that were not detected. The other two squares represent the images that were mismatched in detection.*

A majority of the test have also exhibited a possibility for the networks to overfit or underfit, depending on the performance of the layers. Research from previous papers shows that splitting the initial training data into k subsets and training the model k times can help in preventing such incidents. In each training, it uses one subset as the testing data and the rest as training data. However, prediction tests conducted using conventional methods including Naïve Bayes, KNN, CART(Classification and Regression Trees), LDA(latent Dirichlet Association) show that such means are not powerful enough to reach higher accuracies, even at multiple layering and neurons. Other methods to reduce overfitting and underfitting include:-

1. Holding back a validation dataset from the initial training data to estimate how well the model generalizes on new data.
2. Simplify the model. For example, using fewer layers or less neurons to make the neural network smaller.
3. Use more data and continue to add more batches to both the training and testing sets.
4. Reduce dimensionality in training data such as PCA: it projects training data into a smaller dimension to decrease the model complexity.
5. Stop the training early when the performance on the testing dataset has not improved after a number of training iterations.

The FPGA and CPU are still excellent hardware options for executing full scale object detection and waste tracking projects. Using a uniform kind of dataset is still highly favored as it helps the system learn better and arrive to coefficients for the features faster. The future of waste detection, debris removal and waste tracking will depend highly on the availability of models, well into the reaches of the public. As the simulations in this project have proven, it is possible to build such schemes through open vision and attaching them to well functioning platforms like Python and OpenCV. As the project continues to test on additional datasets with more improved architectures and model algorithms, they can be made better prepared for carrying out field studies in actual water bodies, either through a stationary collection unit or through a mobile unit. Certain model types will, of course, perform better than others and the results in this project provide a gateway of understanding which builds are more feasible. As time goes on, the models will see greater improvement and the push towards innovation in waste debris studies will benefit from the findings in this report and the code developed to supplement the results.